\pdfoutput=1
\documentclass{article}

    \PassOptionsToPackage{numbers, sort, compress}{natbib}

\usepackage[preprint]{neurips_2026}

\usepackage[utf8]{inputenc} 
\usepackage[T1]{fontenc}    
\usepackage{url}            
\usepackage{booktabs}       
\usepackage{amsfonts}       
\usepackage{nicefrac}       
\usepackage{microtype}      
\usepackage{xcolor}         

\usepackage{amsmath}
\definecolor{linkblue}{HTML}{6870ae}
\usepackage[pagebackref=true,colorlinks,citecolor=brown]{hyperref}
\hypersetup{
  urlcolor  = linkblue
}
\usepackage{multirow} 
\usepackage{makecell} 
\usepackage{wrapfig}
\usepackage[inkscape=false]{svg} 
\usepackage{cleveref}
\usepackage{caption}
\usepackage{fontawesome5}

\usepackage{graphicx}
\usepackage{float}
\floatstyle{plaintop}
\restylefloat{table}
\usepackage[export]{adjustbox}

\usepackage{algorithm}
\usepackage{algpseudocode}

\usepackage{amsthm}
\usepackage{subcaption}
\usepackage{etoolbox}
\NewDocumentCommand{\heng}
{ mO{} }{\textcolor{red}{\textsuperscript{\textit{Heng}}\textsf{\textbf{\small[#1]}}}}
\NewDocumentCommand{\chihan}
{ mO{} }{\textcolor{blue}{\textsuperscript{\textit{Chi}}\small[#1]}}  
\NewDocumentCommand{\zhenhailong}
{ mO{} }{\textcolor{purple}{\textsuperscript{\textit{Zhenhailong}}\textsf{\textbf{\small[#1]}}}}

\usepackage{pifont}
\newcommand{\cmark}{\textcolor{green}{\ding{51}}}
\newcommand{\xmark}{\textcolor{red}{\ding{55}}}

\title{Temporal Difference Visual Representation Learning: You Don't Need Strong Assumptions}
\title{Temporal Difference in Vision: You Don't Need Strong Assumptions for Representation Learning}
\title{You Don't Need Strong Assumptions: Visual Representation Learning via Temporal Differences}
\title{You Don't Need Strong Assumptions---Temporal Difference Visual Representation Learning}

\title{You Don't Need Strong Assumptions: Visual Representation Learning via Temporal Differences}

\newcommand{\pa}{TDV} 

\newcommand{\blfootnote}[1]{%
  \begingroup
    \renewcommand\thefootnote{}
    \footnote{#1}%
    \addtocounter{footnote}{-1}
  \endgroup}

\author{%
  \textbf{Ninad Daithankar}\textsuperscript{*1},
  \textbf{Alexi Gladstone}\textsuperscript{*1},
  \textbf{Yann LeCun}\textsuperscript{2},
  \textbf{Heng Ji}\textsuperscript{1} \\
  \textsuperscript{1}UIUC \quad
  \textsuperscript{2}New York University  \quad
  \\
  \textbf{\href{https://temporal-difference-vision.github.io}
     {\faGlobe\enspace{temporal-difference-vision.github.io}}
    \quad
  \href{https://github.com/ninaddaithankar/tdv}
     {\faGithub\enspace{github.com/ninaddaithankar/TDV}}}
}

\begin{document}

\maketitle

\blfootnote{\textsuperscript{*}Equal Contribution. Correspondence to Alexi Gladstone:  \href{mailto:alexigladstone@gmail.com}{\faEnvelope\enspace{alexigladstone@gmail.com}}. Work done while supported as a Flapping Airplanes Fellow.}

\begin{abstract}
\label{sec:abstract}
Progress in AI has largely been driven by methods that assume less. As compute and data increase, approaches with weaker inductive biases generally outperform those with stronger assumptions.
This is particularly characteristic of the field of Visual Representation Learning, where approaches have gone from being dominated by Supervised Learning, to Weakly Supervised Learning, to the now widespread success of Self-Supervised Learning without human labels.
Yet, even modern Self-Supervised Learning approaches still depend on strong inductive biases such as augmentations, masking, or cropping. 
If this trend holds, even these remaining biases should become bottlenecks at scale---and our experiments confirm this: the optimal strength of inductive biases decreases as data grows. This motivates the search for approaches that rely on fewer assumptions.
To this end, we introduce \textbf{Temporal Difference in Vision} (\textbf{{\pa}}), a new 
paradigm 
for self-supervised learning from video that avoids existing inductive biases, relying instead on a causal assumption that the past causes the future. {\pa} functions by jointly training an image encoder and a motion encoder so that the current frame's representation plus the encoded motion equals the next frame's representation.
Despite not leveraging any strong inductive biases, {\pa} matches state-of-the-art recipes on dense spatial tasks, 
laying the foundation
for representation learning without strong assumptions.

\end{abstract}

\section{Introduction}
\label{sec:intro}

Deep learning has achieved remarkable progress over the last decade and a half, advancing from simple object classification~\cite{krizhevsky2012imagenet} to high-resolution image generation~\cite{rombach2022high} and sophisticated cross-modal reasoning~\cite{openai2023gpt4}. This progress has largely been driven by methods that more effectively leverage increasing data and computation~\cite{kaplan2020scaling, hoffmann2022training}, where approaches with weaker \textit{inductive biases}\footnote{We broadly use the term inductive biases and assumptions interchangeably in this work.} tend to outperform those with stronger assumptions as scale increases~\cite{sutton2019bitter, dosovitskiy2020image, oquab2023dinov2, simeoni2025dinov3, moutakanni2024you, chung2025shaping, gladstone2025energy}.

This principle is illustrated by the evolution of visual representation learning, where progress has largely been driven by approaches using progressively weaker inductive biases. 
For example, early supervised learning with convolutional neural networks (CNNs)~\cite{krizhevsky2012imagenet, he2016deep} assumed that human-annotated labels captured the semantic structure of images, while convolutional architectures imposed spatial locality biases on representations. Moving away from labels, self-supervised contrastive approaches such as SimCLR~\cite{chen2020simple} and MoCo~\cite{he2020momentum} instead pulled augmented views of the same image together and pushed different images apart, but made strong assumptions about the distances between negative pairs. To address this flaw, self-distillation approaches~\cite{grill2020bootstrap} relaxed these assumptions via a slow-moving teacher, and the subsequent adoption of Vision Transformers (ViTs)~\cite{dosovitskiy2020image, caron2021emerging} discarded the locality and translation equivariance biases of CNNs in favor of global attention. Now, modern approaches combining self-distillation with ViTs achieve state-of-the-art performance~\cite{oquab2023dinov2,simeoni2025dinov3}.

\begin{figure}[t]
    \centering
    \includegraphics[width=0.9\linewidth]{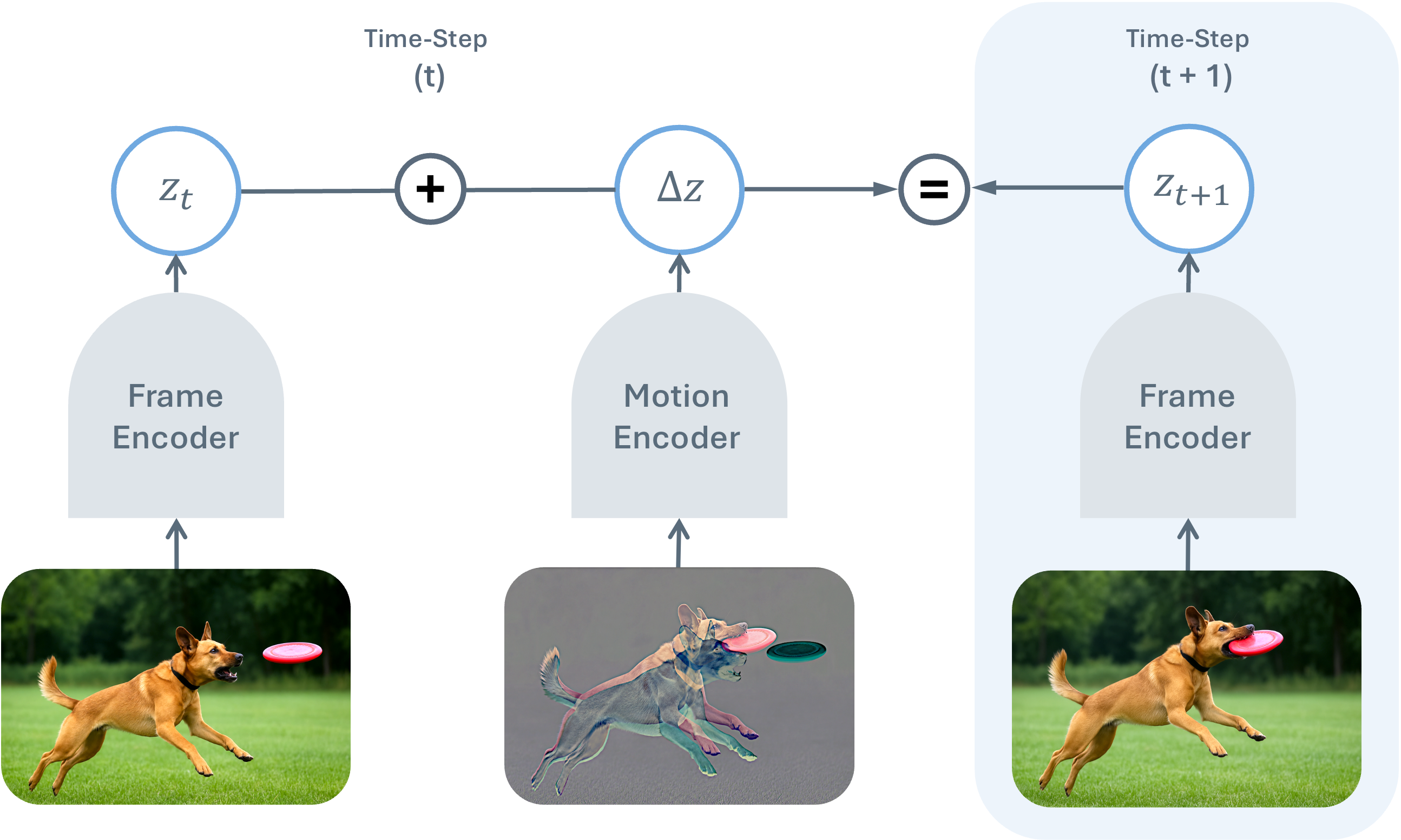}
    \caption{\textbf{{\pa} Frame and Motion Encoding Intuition.} {\pa} learns to encode frames such that the current frame's representation, when added to a learned motion encoding, predicts the next frame's representation. Because video has high temporal consistency, the raw RGB pixel difference between frames is intrinsically lower rank than the frames themselves, shown here as the edge outlines of a dog and a frisbee. The motion encoder compresses these high-dimensional RGB differences into abstract motion-level features.
    }
    \label{fig:tdv-intuition}
\end{figure}

Interestingly, this pattern of weaker assumptions leading to better performance mirrors biological evolution, where innate instincts play a role analogous to hardcoded inductive biases. Across animals, more capable species tend to hardcode less behavior into their genome---insects rely heavily on innate behavioral programs, while mammals depend substantially more on learned behavior~\cite{tinbergen2020study, dukas2008evolutionary}. This trend is even more distinct in primates, and most pronounced in humans, who rely heavily on learning from experience rather than hardcoded behavior~\cite{reader2002social, gomez2024evolution}. In both the case of visual representation learning as well as hardcoded behavior for biological intelligence, less hardcoded structure enables greater asymptotic performance given sufficient scale.

To further support this principle, we empirically test how the optimal strength of inductive biases changes with data scale (Figure~\ref{fig:philosophical_assumption_exps}). We find that as data scale increases, weaker inductive biases outperform stronger ones asymptotically---reinforcing that minimizing assumptions becomes increasingly important as scale increases.

Motivated by this trend, we argue for a new approach to visual representation learning that avoids the inductive biases relied upon by existing methods (we discuss these biases in Section~\ref{sec:additional_intuition}). Removing them naively, however, leaves no learning signal and collapses the representation (Table~\ref{tab:dino_ablation_augmentation}).

Therefore, a natural question emerges---\textit{``What assumptions should our model have, if not reliant on existing inductive biases?''} We argue for assuming \textit{causality}: that causes precede their effects, and the immediate future is therefore predictable from the past. This principle is foundational across physics, from classical mechanics to relativistic field theories~\cite{einstein1905electrodynamics} to modern formulations of quantum theory~\cite{d2018causality}.\footnote{We do not claim the stronger thesis of determinism---that the past \textit{uniquely} determines the future---which is incompatible with quantum mechanics under standard assumptions~\cite{bell1964einstein}. We assume only the weaker principle, that the immediate future is generally predictable from the past, sufficient to provide a learning signal.} 
Unlike existing inductive biases for representation learning, we argue causality is weak, and
domain agnostic.
Additionally, because causality is inherently temporal, applying it points towards training over video, rather than images.\footnote{Learning image encoders from video departs from common practice in representation learning, which historically trains them on image datasets.}

\begin{figure} [t]
    \centering
    \includegraphics[width=0.8\linewidth]{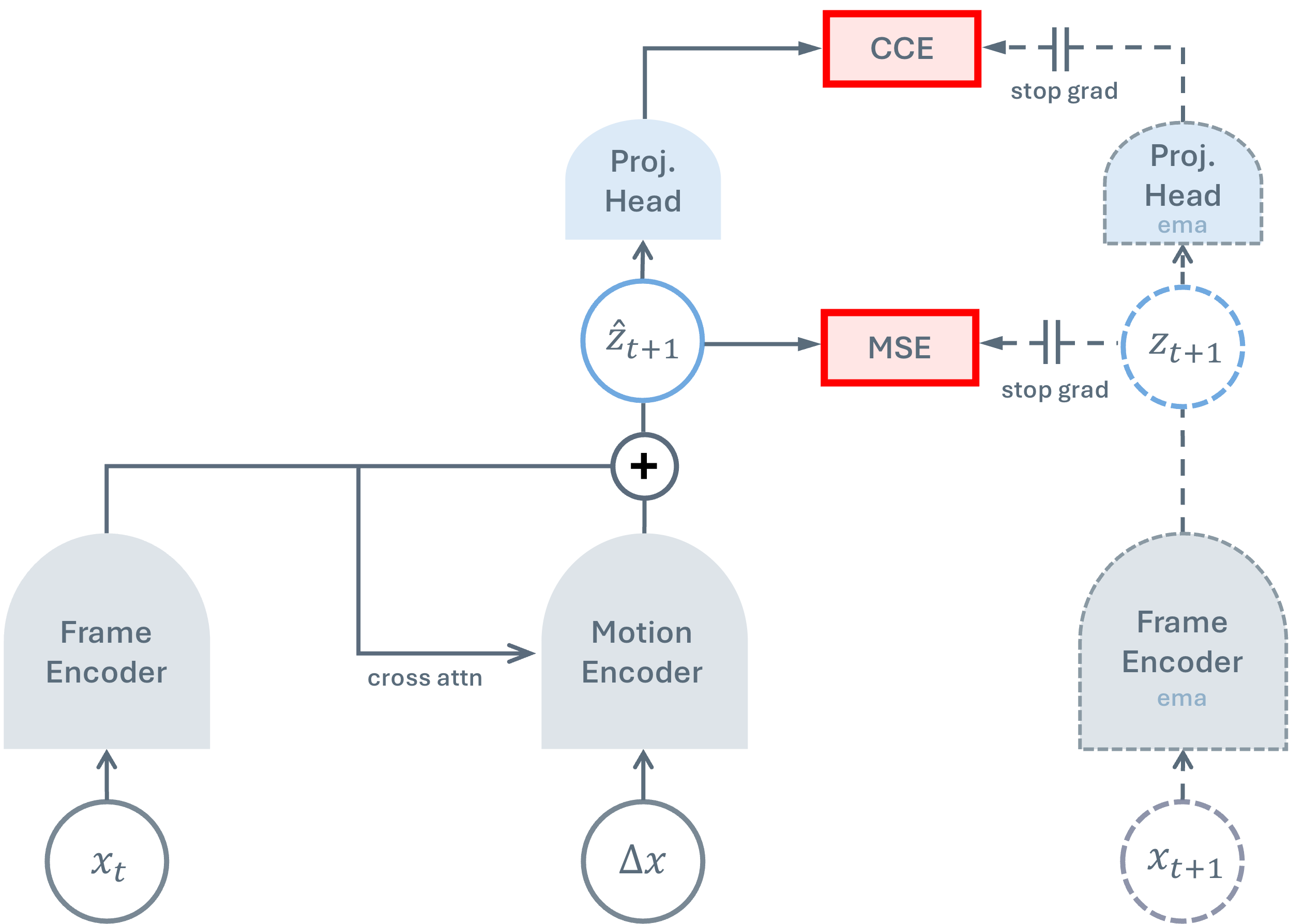}
    \caption{\textbf{{\pa} Architecture.} 
    {\pa} predicts the next frame's representation by adding a learned motion vector to the current frame's representation. \textit{Left (student):} the frame encoder embeds the current frame, while the motion encoder turns the raw pixel difference between frames into a latent motion shift, conditioned on the current frame via cross-attention. Their sum is the predicted representation of the next frame. \textit{Right (teacher):} an EMA copy of the frame encoder embeds the true next frame to supply the target. Two losses act on the prediction: a mean-squared error on the representations enforces the causal next-frame constraint, and a DINO-style~\cite{caron2021emerging} cross-entropy on the projection heads prevents collapse. Stop-gradients block the teacher from receiving gradients. Figure style inspired by \cite{bardes2023v, assran2025v}.
    }
    \label{fig:tdv-full-architecture}
\end{figure}

To achieve this, we jointly train a frame encoder and a motion encoder so that, given two consecutive frames, the embedding of the current frame plus the embedding of the frame delta matches the embedding of the next frame (visualized in Figure~\ref{fig:tdv-intuition}). 
Because consecutive frames are close in time, and video has high temporal consistency, the frame delta is intrinsically low-rank, encouraging the motion encoder to capture compact spatial change rather than full scene appearance. 
Given its similarity to Temporal Difference in Reinforcement Learning~\cite{sutton1988learning}, we call our approach \textbf{Temporal Difference in Vision ({\pa})}. {\pa} naturally enables learning without restrictive inductive biases or modality-dependent assumptions. Empirical results demonstrate {\pa} is able to learn dense spatial features comparable to state-of-the-art visual encoders such as DINO~\cite{caron2021emerging} and iBOT~\cite{zhou2021ibot} without relying on strong assumptions during pretraining.

Our contributions are as follows:

\begin{itemize}
\item We confirm our hypothesis regarding the importance of weaker assumptions as scale increases through controlled experiments, further motivating {\pa}.
\item We present {\pa}, a new
paradigm 
for learning visual representations that avoids the strong inductive biases of existing approaches.
\item We demonstrate promising empirical performance for {\pa}, achieving dense spatial features on par with modern approaches that leverage stronger inductive biases.
\end{itemize}

\section{Related Work and Background}

\subsection{Self-Supervised Representation Learning}
Self-supervised representation learning has the goal of learning representations without any labels. Early work in this domain learned representations primarily via autoencoding~\cite{rolfe2013discriminative, vincent2008extracting}, where models were trained to reconstruct inputs directly in pixel space. Over time, the field has shifted primarily from raw pixel space reconstruction towards Joint Embedding Predictive Architectures (JEPAs)~\cite{lecun2022path}, where prediction is done in a latent space as opposed to in the raw pixel space. Such models abstract away irrelevant, unpredictable information, such as background pixels in a scene, in favor of modeling more important information---a form of learning with weaker inductive biases. Recent empirical~\cite{simeoni2025dinov3, assran2025v} and theoretical~\cite{balestriero2024learning, littwin2024jepa} evidence reinforces the benefits of JEPAs, where raw pixel space reconstruction is often theoretically predicted and empirically observed to produce less informative features for perception~\cite{balestriero2024learning}.

The progression within the JEPA family illustrates a recurring pattern: methods with weaker inductive biases have steadily displaced those with stronger ones. Early JEPA approaches relied on contrastive objectives~\cite{chen2020simple, he2020momentum}, which prevent collapse by pushing apart representations of distinct images while pulling together augmented views of the same image. However, contrastive methods impose a strong relational prior—that randomly sampled images should be dissimilar in representation space---which is only approximately correct, since sampled pairs frequently depict semantically related content. They also depend on large batches of negative samples, limiting scalability~\cite{radford2021learning}. Self-distillation approaches~\cite{grill2020bootstrap, caron2021emerging} relax this prior entirely by replacing negative pairs with a slow-moving teacher network: the student is trained to match the teacher's output on a different view of the same image, while the teacher is updated as an exponential moving average of the student. This eliminates the negative relational assumption between images, with centering and stop-gradient mechanisms preventing trivial collapse. Modern state-of-the-art vision foundation models such as DINOV3~\cite{simeoni2025dinov3} and V-JEPA2~\cite{assran2025v} build on this self-distillation paradigm,\footnote{Some of these architectures, such as DINOV3, are technically considered Joint-Embedding Architecture (JEA) variants, due to not conditioning on a latent variable $z$.} reflecting the broader trajectory of the field toward progressively weaker inductive biases.

\subsection{Temporal Difference for Representation Learning}
Despite this trajectory of weakening inductive biases, modern self-distillation approaches still rely on image-level inductive biases such as cropping, masking, or augmentations. A natural alternative is to source paired views from time itself, using temporally adjacent frames in a video. Several works explore this direction~\cite{simonyan2014two, oord2018representation, han2019video, recasens2021broaden, bardes2023mc}. \citet{feng2023mutual} train supervised models over temporal difference features, minimizing mutual information to disentangle task-relevant motion from noise. \citet{wang2021tdn} model low-level frame deltas for action recognition, but rely on a global channel attention mechanism to recalibrate features across long-range differences. \citet{maes2026leworldmodel} predict future frames in latent space, but target world modeling rather than transferable representations. Most closely related to {\pa}, Midway Networks~\cite{hoang2025midway} learn representations directly from temporal differences in video, adding an invariance objective over cropped patches to target semantic performance. In each case, the temporal signal is paired with an additional inductive bias---supervision, attention recalibration, a world-modeling objective, or augmentation-based invariance. With {\pa}, we instead focus on learning from temporal difference alone, without any such biases.

\section{{\pa} Approach}

\subsection{{\pa} Intuition}

Learning representations without strong inductive biases is notoriously challenging~\cite{chen2020simple, bardes2022vicreg, zbontar2021barlow}; removing assumptions such as augmentations or masking often leads to degraded representations or collapse. We confirm this by removing key inductive biases in the well-known DINO~\cite{caron2021emerging} recipe, observing poor performance and eventual collapse (Table~\ref{tab:dino_ablation_augmentation}). These results, and our deeper motivation to remove inductive biases, raise a natural question: \textit{``What is the weakest assumption that still provides sufficient signal to avoid collapse?''}

Answering this requires understanding why assumptions hurt performance in the first place. We argue that assumptions encode beliefs that are only approximately correct~\cite{ericsson2021self, xiao2020should}, and at scale these approximations restrict what can be learned~\cite{sutton2019bitter, dosovitskiy2020image, moutakanni2024you}.
Instead, an assumption that is exactly, rather than approximately, correct, would impose no such bottleneck, providing a learning signal without restricting what can ultimately be learned.\footnote{By ``exactly correct'' we refer to a property of the learning objective, not a claim about causality itself: next-frame prediction imposes no invariance constraint, and therefore never requires the encoder to discard a factor of variation. Augmentation- and masking-based objectives instead enforce invariance to a chosen transformation, discarding the corresponding information by construction, which can degrade downstream performance on tasks where that information is needed (Section~\ref{sec:additional_intuition}).}

\begin{wraptable}{r}{0.62\textwidth}
\centering
\caption{\textbf{Removing DINO's Inductive Biases Degrades Performance.} 
Progressively removing DINO's augmentations for pre-training on SSV2 degrades KNN performance, and eventually causes representation collapse. {\pa}, by contrast, avoids collapse without these inductive biases.
2G and 8L denote 2 random global and 8 random local crops respectively. Full augmentations includes random flip, color jitter, Gaussian blur, and solarization.}
\label{tab:dino_ablation_augmentation}
\small
\begin{tabular}{lccc}
\toprule
Setup & $\uparrow$Top-1 & $\uparrow$Top-5 & Avoids Collapse \\
\midrule
2G + 8L, full aug   & 24.63 & 40.19 & {\color{green}\checkmark} \\
     - augmentations& 16.44 & 26.82 & {\color{green}\checkmark} \\
     - local crops& 13.64 & 23.07 & {\color{green}\checkmark} \\
- random crop on 1G  & 8.04  & 14.91 & {\color{green}\checkmark} \\
- random crop on both G    & 0.84  & 2.37  & {\color{red}\texttimes} \\
\midrule
Full TDV Recipe (ours)                & 8.79  & 17.05  & {\color{green}\checkmark} \\
\bottomrule
\end{tabular}
\end{wraptable}

We argue such an assumption exists: \textit{causality}---the principle that the past is predictive of the future. This principle is foundational across physics~\cite{einstein1905electrodynamics, d2018causality}, with classical mechanics serving as a canonical example---an object's position, velocity, and acceleration are sufficient to predict its trajectory. Unlike assumptions such as ``augmented views should be invariant'' or ``masked and unmasked images should be similar,'' causality is domain-agnostic and, we argue, exactly rather than approximately correct. Perhaps this assumption is part of the reason autoregressive Large-Language Models have been so successful~\cite{openai2023gpt4, touvron2023llama, guo2025deepseek}, as they assume causality, which reflects the data-generating procedure of language.

Leveraging causality, however, is non-trivial. Image representations are traditionally learned from static image datasets~\cite{oquab2023dinov2, caron2021emerging, bardes2022vicreg}, which lack the temporal dimension causality requires.\footnote{One could apply causality within a single image by predicting one patch from another~\cite{van2016pixel}; however, this does not reflect causality as it operates in the world, where causes precede effects in \textit{time}.} We therefore argue for learning image representations from \textit{video}, where consecutive frames provide the temporal structure causality demands.

Specifically, we train an image encoder jointly with a motion encoder such that the image encoder's representation of a frame, added to the motion encoder's representation of the change between frames, yields the image encoder's representation of the next frame (visualized in Figure \ref{fig:tdv-intuition}). Intuitively, this motion representation generally has low intrinsic rank, since the semantic change between consecutive frames is typically small. By analogy to Temporal Difference in Reinforcement Learning~\cite{sutton1988learning}, we call this approach \textbf{Temporal Difference in Vision ({\pa})} (Figure~\ref{fig:tdv-full-architecture}).

Beyond its motivation from causality, {\pa} can also be viewed as a form of self-distillation~\cite{grill2020bootstrap, oquab2023dinov2, balestriero2023cookbook}. However, rather than forcing invariance across hand-crafted augmentations such as cropping, rotating, or masking, the ``augmentation'' is induced by time, with temporally consecutive frames serving as the two views. The changes produced by this temporal augmentation are then modeled explicitly by the motion encoder, rather than being discarded via an invariance objective. 
This can be viewed as a learned, latent-space analog of the motion vectors used in classical video codecs~\cite{wiegand2003overview}, which similarly represent video as a frame plus the motion to the next.
Intuitively, this objective forces {\pa}'s representations to be sufficiently informative of the current frame as well as rich enough to predict the next frame.
%


\subsection{{\pa} Architecture}
\label{sec:architecture}

Having established causality as our guiding assumption, we now derive the architecture for {\pa}. Our goal is to follow a simple principle: \emph{the representation of a frame, combined with the change that occurs between frames, should yield the representation of the next frame.}

\paragraph{Learning a representation space.} Following our principle, we first need 
a way to map the frames from RGB space into a meaningful representation space. We 
therefore learn a \textbf{frame encoder} $f_\theta$ that maps each frame $x_t$ to 
a sequence of token embeddings:
\begin{equation}
    z_t = f_\theta(x_t) \in \mathbb{R}^{n \times D},
\end{equation}
where $z_t$ is the representation for frame $x_t$, $n$ is the number of spatial 
patches plus an additional \texttt{[CLS]} token, and $D$ is the embedding dimension. 
Our causal principle then becomes a constraint in this embedding space: the change 
between frames, when encoded appropriately, should be sufficient to predict the next 
frame's embedding.

\paragraph{Encoding change in representation space.} The raw RGB difference $\Delta x_t = 
x_{t+1} - x_t$ captures what changed in pixel space, but we need to map this into a corresponding shift $\Delta z_t$ in the latent space. Importantly, $\Delta x_t$ is intrinsically lower rank than the frames themselves, as the background scene pixels remain largely unchanged between adjacent frames, and only moving regions contribute a non-zero signal (as visualized in Figure~\ref{fig:tdv-intuition}). We therefore learn a 
\textbf{motion encoder} $m_\phi$ that takes the change in RGB space $\Delta x_t$, and predicts the change in representation space $\Delta z_t$. Since the same pixel-level change can carry different semantic meanings depending on visual context, we condition the motion encoder on the current frame's embedding $z_t$ via cross-attention, grounding the motion prediction in the semantic state of the current frame:
\begin{equation}
    \Delta z_t = m_\phi(\Delta x_t;\, z_t)
\end{equation}
\paragraph{Additive latent composition.} With the frame encoder learning to encode the current frame in representation space and the motion encoder learning the change in representation space, predicting the next frame's representation reduces to a simple additive composition:
\begin{equation}
    \hat{z}_{t+1} = z_t + \Delta z_t
\end{equation}
This decomposition of $\hat{z}_{t+1}$ into $z_t$ and $\Delta z_t$ cleanly separates the goal into two objectives: the frame encoder is responsible for learning the content in a frame, and the motion encoder learns how that content evolves over time. 

\begin{wrapfigure}{r}{0.5\textwidth}
    \centering
    \includesvg[width=.5\textwidth]{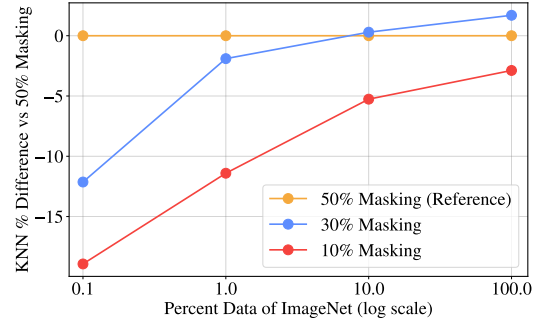}
    \caption{\textbf{Need for Assumptions Decreases as Data Scale Increases.} 
    KNN accuracy on ImageNet-1k for three masking ratios (our proxy for assumption strength) across data scales, reported as percentage-point difference vs.\ 50\% masking. At 0.1\% data, 30\% masking trails 50\% masking by over 12 percentage points; by 100\% data, it surpasses 50\% masking. Lighter masking (10\%) follows the same trend but lags, suggesting it may eventually surpass other masking ratios with increased scale. 
    These results demonstrate that the optimal amount of inductive bias decreases with scale, motivating {\pa}. 
    }
    \label{fig:philosophical_assumption_exps}
\end{wrapfigure}

\paragraph{Preventing collapse.} Supervising the next frame's predicted representation $\hat{z}_{t+1}$ requires a target: $z_{t+1}$, the next frame $x_{t+1}$ encoded by the frame encoder. However, this makes the {\pa} recipe prone to collapse, as $z_{t+1}$ is also produced by the same frame encoder that is being trained. Therefore, the encoder can trivially achieve near-zero loss by collapsing all representations to a constant, making the target easy to predict, but meaningless~\cite{balestriero2023cookbook}. To prevent this, we adopt a teacher-student framework following DINO~\cite{caron2021emerging}: we maintain two copies of the frame encoder, a \emph{student} updated by gradient descent, and a \emph{teacher}, whose parameters are a slowly-evolving exponential moving average (EMA) of the student. The target is then produced by the teacher, which we denote as $z^{\text{teacher}}_{t+1}$. Both the student and teacher pass their representations through respective projection heads, and we apply a cross-entropy loss between the resulting prototype distributions. This penalizes collapse directly: if all frames map to the same representation, the distributions become identical across frames and the cross-entropy loss increases, forcing the encoder to maintain discriminative representations. The teacher's parameters evolve slowly via EMA, ensuring the student and teacher 
remain sufficiently different at any point in training to provide stable, non-trivial, prediction targets that the student cannot trivially satisfy by collapsing to the same distribution~\cite{grill2020bootstrap, caron2021emerging}. We illustrate the complete architecture in Figure~\ref{fig:tdv-full-architecture}.


\subsection{{\pa} Training Objective}

With the architecture set, we now describe how each component is supervised. 
{\pa} is trained with a weighted combination of two losses, each targeting a distinct 
objective.

\paragraph{Temporal prediction loss ($\mathcal{L}_{\text{mse}}$).} The first loss directly supervises our causal 
principle established in Section~\ref{sec:architecture}: the motion encoder must 
produce a $\Delta z_t$ that, when added to $z_t$, accurately recovers the next 
frame's embedding. This is enforced via a mean-squared error between the predicted 
next-frame embedding $\hat{z}_{t+1} = z_t + \Delta z_t$ and the teacher-encoded 
target $z^{\text{teacher}}_{t+1}$:
\begin{equation}
    \mathcal{L}_{\text{mse}} = \left\| \hat{z}_{t+1} - \text{sg}(z^{\text{teacher}}_{t+1}) \right\|_2^2,
\end{equation}
where $\text{sg}(\cdot)$ denotes stop-gradient, ensuring this loss only updates 
the motion encoder and student frame encoder, not the teacher.

\paragraph{Self-distillation loss ($\mathcal{L}_{\text{dino}}$).} The second loss addresses the collapse problem 
described in Section~\ref{sec:architecture}: without an additional signal, the frame 
encoder can trivially satisfy $\mathcal{L}_{\text{mse}}$ by collapsing all 
representations to the same embedding. We therefore apply a cross-entropy objective inspired by 
DINO~\cite{caron2021emerging} between student and teacher projection distributions, 
with one extension: we apply this loss over \emph{both} the \texttt{[CLS]} token 
and \emph{the patch tokens}, encouraging spatially consistent representations at 
the patch level beyond what the original DINO formulation provides. Let $p_s$ and 
$p_t$ denote the student and teacher projection distributions, normalized with 
temperatures $\tau_s$ and $\tau_t$ respectively (in practice, we set $\tau_t = 
\tau_s = 0.1$). The loss is then:
\begin{equation}
    \mathcal{L}_{\text{dino}} = -\sum_{k} p_t^{(k)} \log p_s^{(k)},
\end{equation}
where $k$ indexes over the $K$ prototype dimensions of the projection head. The teacher distribution is additionally centered with a running mean to prevent 
dimensional collapse in the absence of temperature asymmetry~\cite{caron2021emerging}.

Putting these together, we get the complete training objective for {\pa}:
\begin{equation}
    \mathcal{L} = \lambda_{\text{mse}}\,\mathcal{L}_{\text{mse}} + \lambda_{\text{dino}}\,\mathcal{L}_{\text{dino}},
\end{equation}
where $\lambda_{\text{mse}}$ and $\lambda_{\text{dino}}$ are tunable hyperparameters.

\section{Experimentation}
\label{sec:experimentation}

\begin{table}
\centering
\caption{\textbf{Semantic Segmentation Performance With UperNet.} We benchmark the Semantic Segmentation performance of {\pa} compared to iBOT and DINO on ADE20K and Cityscapes. TDV achieves competitive performance relative to iBOT and DINO despite learning without strong inductive biases.}

\label{tab:downstream_semseg}
\small
\begin{tabular}{lcc|cc|cc}
\toprule
 & & & \multicolumn{4}{c}{\textbf{Semantic Segmentation (UperNet)}} \\
 & & & \multicolumn{2}{c|}{ADE20K} & \multicolumn{2}{c}{Cityscapes} \\
Method & Arch & Pretrain & $\uparrow$mIoU & $\uparrow$mAcc & $\uparrow$mIoU & $\uparrow$mAcc \\
\midrule
iBOT~\cite{zhou2021ibot}       & ViT-S & SSv2 & 10.60 & 14.53 & 39.34 & 45.36 \\
DINO~\cite{caron2021emerging}  & ViT-S & SSv2 & \textbf{10.71} & \textbf{14.64} & \textbf{39.85} & \textbf{45.68} \\
TDV                            & ViT-S & SSv2 & 10.54 & 14.48 & 37.54 & 43.09 \\
\midrule
iBOT~\cite{zhou2021ibot}       & ViT-B & SSv2 & 9.94 & \textbf{13.65} & 38.94 & \textbf{44.31} \\
DINO~\cite{caron2021emerging}  & ViT-B & SSv2 & \textbf{10.48} & 11.14 & \textbf{39.97} & 43.09 \\
TDV                            & ViT-B & SSv2 & 9.57 & 10.70 & 36.21 & 42.59 \\
\bottomrule
\end{tabular}
\end{table}

\subsection{Motivating Weaker Assumptions}
\label{sec:motivating_weaker_assumptions}

To provide empirical weight to our philosophical argument---that weaker assumptions yield superior asymptotic performance as data scales---we evaluate various models across different subsets of ImageNet-1k~\cite{russakovsky2015imagenet}. By identifying the top-performing inductive biases at each data scale, we can observe how the optimal ``strength'' of assumptions shifts. Specifically, we conduct these evaluations using data subsets of $0.1\%$, $1\%$, $10\%$, and $100\%$. To measure the strength of these assumptions, we utilize masking with values of $10\%$, $30\%$, and $50\%$ as a continuous proxy (note that these are values not for {\pa}, but for testing our argument regarding weaker assumptions; more details are in Section~\ref{sec:experimental_details}). We use masking for two primary reasons: it allows for a granular axis, unlike discrete changes such as switching from contrastive learning to self-distillation, and it represents a clear spectrum of assumptions. For instance, requiring models to treat images with $50\%$ masking as ``similar'' to their original imposes a strong assumption that only the high-level semantics remaining are sufficient for representation. Alternatively, requiring image similarity at $10\%$ masking is a relatively weak assumption, as most of the image details remain intact.

The results for these experiments are shown in Figure~\ref{fig:philosophical_assumption_exps}, where the results demonstrate a strong trend between which approaches perform best and the amount of data being leveraged. With just $0.1\%$ of ImageNet, the best performing masking ratio is $50\%$, with $30\%$ and $10\%$ masking falling behind by a significant margin. However, as the amount of data increases, $30\%$ masking eventually outperforms $50\%$ masking, with $10\%$ masking approaching the performance of $50\%$ masking. These results demonstrate that as data increases, the optimal amount of assumptions made, represented here as masking ratio, decreases. This further reinforces our motivation for {\pa}---to learn representations without any strong inductive biases. 


\subsection{Downstream Evaluations}

\begin{table}[t]
\centering
\caption{\textbf{Optical Flow and Stereo Depth Evaluation.} {\pa} outperforms iBOT and DINO on most optical flow and stereo depth comparisons, with a small trade-off on stereo depth average error.}
\label{tab:downstream_flow_stereo}
\small
\begin{tabular}{lcc|cc|ccc}
\toprule
 & & & \multicolumn{2}{c|}{\textbf{Optical Flow}} & \multicolumn{3}{c}{\textbf{Stereo Depth}} \\
 & & & \multicolumn{2}{c|}{MPI-Sintel} & \multicolumn{3}{c}{SceneFlow (final)} \\
Method & Arch & Pretrain
  & $\downarrow$EPE (clean) & $\downarrow$EPE (final)
  & $\downarrow$Avg Err. & $\downarrow$bad@0.5px & $\downarrow$bad@1px \\
\midrule
iBOT~\cite{zhou2021ibot}       & ViT-S & SSv2 & 11.31 & 11.27 & \textbf{3.50} & 65.51 & 44.91 \\
DINO~\cite{caron2021emerging}  & ViT-S & SSv2 & 13.03 & 12.92 & 3.64 & 63.25 & 45.30 \\
TDV                            & ViT-S & SSv2 & \textbf{9.84} & \textbf{10.75} & 4.25 & \textbf{56.89} & \textbf{39.70} \\
\midrule
iBOT~\cite{zhou2021ibot}       & ViT-B & SSv2 & 11.66 & 11.82 & \textbf{3.75} & 62.49 & 44.18 \\
DINO~\cite{caron2021emerging}  & ViT-B & SSv2 & 11.63 & \textbf{11.28} & 3.91 & 62.97 & 44.64 \\
TDV                            & ViT-B & SSv2 & \textbf{10.97} & 11.85 & 3.98 & \textbf{54.62} & \textbf{37.33} \\
\bottomrule
\end{tabular}
\end{table}

Following~\cite{carreira2024scaling}, we argue that semantic benchmarks such as linear probing, $k$-NN retrieval, and action recognition probe \textit{ventral stream}~\cite{felleman1991distributed, goodale1992separate} skills (what), and generally do not accurately measure spatial or temporal representation quality. However, understanding structure and motion, which is performed in the human brain by the \textit{dorsal stream}, is fundamental to real-world vision applications such as robotics, autonomous driving, and 3D scene understanding. These tasks are often bottlenecked less by semantic representations and more by low-level spatial-temporal information.
We therefore focus our evaluations on such properties, specifically segmentation, optical flow and stereo depth, as they demand representations to retain spatial structure and temporal correspondence, precisely the properties suppressed by strong semantic priors but preserved by {\pa}.

For our experimental setup, we pretrain all models on the SomethingSomethingV2 (SSV2)~\cite{goyal2017something} dataset, as it has well-defined motion data and is a standard video benchmark~\cite{bardes2024revisiting}.
We then evaluate all models on the downstream tasks using the pretrained backbones, with individual setup details provided in Appendix~\ref{sec:downstream_eval_details}.

\begin{figure}[t]
    \centering \small
    \includegraphics[width=1\linewidth]{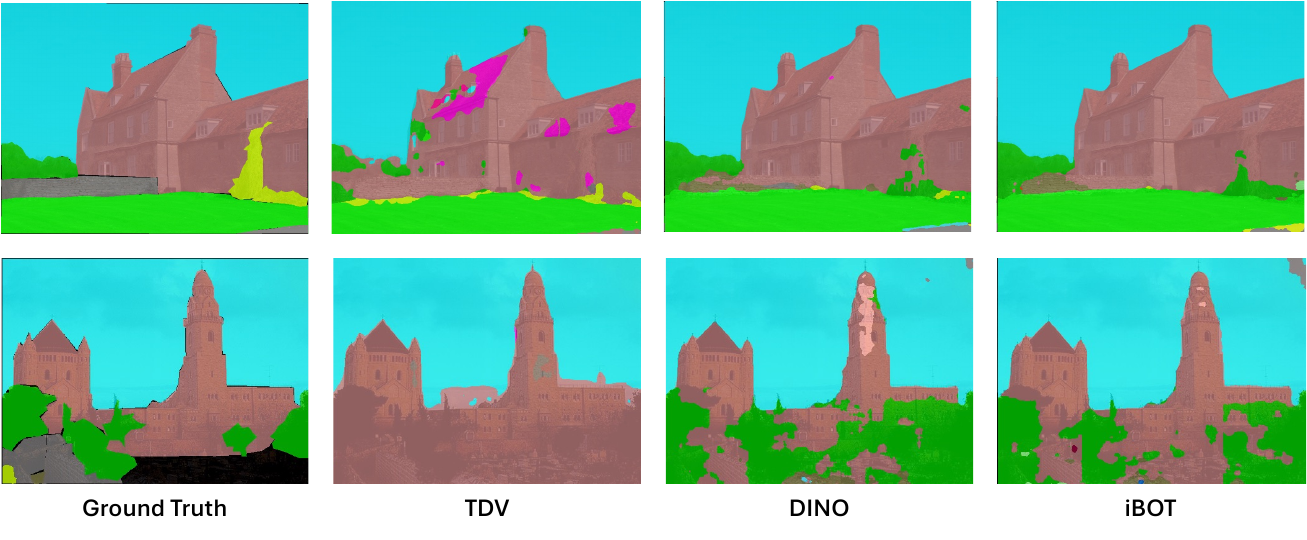}
        \caption{\textbf{Semantic Segmentation on ADE-20K (UperNet).}  TDV performs competitively to DINO and iBOT, with broader region extents but slightly worse boundary separation.}

    \label{fig:semantic_segmentation-viz}
\end{figure}

On semantic segmentation, {\pa} achieves results comparable to DINO and iBOT, trailing behind by a small margin on both mIoU (mean intersection over union) and mAcc (mean per-class accuracy) as shown in Table \ref{tab:downstream_semseg}. The competitiveness, visualized in Figure \ref{fig:semantic_segmentation-viz}, suggests that {\pa} is capable of learning spatially coherent features that a segmentation head can leverage even without an explicit semantic objective. This competitiveness holds despite TDV producing less object-focused \texttt{[CLS]}-token attention than DINO and iBOT (Figure~\ref{fig:attention_visualization}), likely because segmentation relies on patch-level rather than \texttt{[CLS]}-token features. 
We believe the remaining performance gap likely reflects the absence of augmentations like local cropping, which may provide better semantic context.

\begin{figure}[t]
    \centering \small
    \includegraphics[width=1\linewidth]{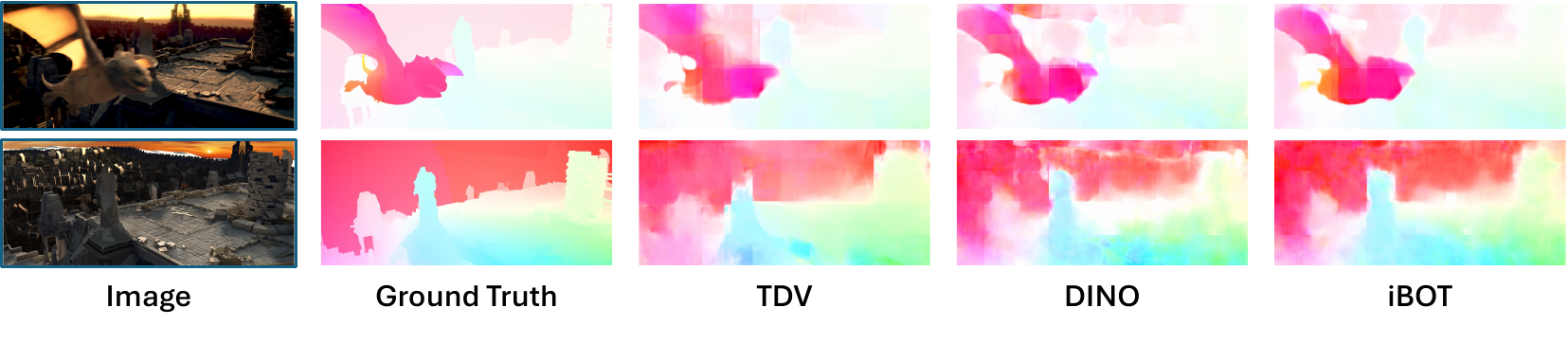}
        \caption{\textbf{Optical Flow on MPI-Sintel.} TDV produces locally consistent flow compared to DINO and iBOT, though artifacts remain in occluded regions across all methods.}

    \label{fig:optical-flow-viz}
\end{figure}

We evaluate {\pa} on temporal tasks against DINO and iBOT in Table~\ref{tab:downstream_flow_stereo}.
On optical flow, {\pa} consistently outperforms both DINO and iBOT on EPE (endpoint error, the average pixel-level distance between predicted and ground truth flow vectors). We believe this can be attributed to {\pa} explicitly learning to predict how representations evolve between frames, which naturally preserves the local motion structure that methods trained on images with invariance augmentation tend to discard (optical flow predictions are visualized in Figure~\ref{fig:optical-flow-viz}).
On stereo depth, {\pa} achieves lower ``bad'' pixel rates at both the 0.5px and 1px thresholds across both architectures, indicating that {\pa} makes significantly fewer large correspondence errors than DINO and iBOT. The slightly higher average disparity error suggests that while {\pa} makes fewer large mistakes, it can still struggle to recover precise depth in ambiguous regions where semantic context would otherwise help. 
These performance gains carry over to the features themselves: Figure~\ref{fig:pca-vis} shows PCA visualizations of patch-level features, where {\pa} produces spatially coherent feature maps.

\subsection{{\pa} Ablation Studies}

We ablate the key design choices of {\pa} to identify the components critical to performance and stability. We pretrain {\pa} on SSv2 and use online ImageNet KNN Top-5 accuracy~\cite{oquab2023dinov2} as a proxy for general performance, as it is cheap to compute
during training, and gives a meaningful signal for representation quality. We show ablation results in Table~\ref{tab:tdv_ablations}.

The two ablations that cause training to collapse the most are removing the motion encoder and removing the MSE loss, highlighting them as two critical components of {\pa}. The motion encoder provides the temporal signal 
necessary for learning, while the MSE loss directly supervises it to predict meaningful 
changes in representation space. Notably, removing the motion encoder entirely and 
relying solely on the DINO loss across consecutive frames---effectively reducing {\pa} 
to a simple temporal invariance objective---is also not sufficient to learn representations, suggesting that explicitly modeling temporal differences between frames is a necessary choice.

Among the remaining design choices, including the \texttt{[CLS]} token in cross-attention and applying the DINO loss on the \texttt{[CLS]} token both contribute meaningfully to performance.
These suggest that grounding motion predictions in a global scene representation helps the motion encoder focus on semantically meaningful changes. 
For the teacher's output distribution, removing centering causes a much larger performance drop than removing temperature sharpening, which we attribute to centering preventing the distribution from becoming too peaked on a single mode, a subtler form of collapse.
Finally, we find that standard absolute positional encodings consistently outperform RoPE~\cite{su2024roformer} across our experiments. While this ablation study addresses the components that work, we document all the other design choices and training strategies we tried that did not work in Appendix \ref{sec:negative_results}.

\begin{table}[t]
    \centering \small
    \caption{\textbf{Performance Impact of Key TDV Ablations.} We ablate key components of the full TDV recipe pretrained on SSv2 and report KNN (Top-5) accuracy on ImageNet-1k alongside training stability. We find that removing the motion encoder or MSE loss causes training collapse, identifying them as critical components. Similarly, centering, DINO loss on the \texttt{[CLS]} token, and cross-attention with the \texttt{[CLS]} token contribute meaningfully to performance without affecting stability.}
    \begin{tabular}{l c c}
        \toprule
         & KNN (Top-5) $\uparrow$ & Avoids Collapse\\
        \midrule
        Full {\pa} recipe & 17.05         & \cmark\\
        No Temperature & 15.85          & \cmark\\
        No Centering & 11.15         & \cmark\\
        No [CLS] in Cross Attention & 10.78 & \cmark\\
        No Centering or Sharpening & 10.68          & \cmark\\
        No DINO Loss on [CLS]& 10.66          & \cmark\\
        RoPE instead of Positional Enc. & 10.25         & \cmark\\
        No Motion Encoder & 1.87          & \xmark\\
        No MSE Loss & 1.58          & \xmark\\
        \bottomrule
    \end{tabular}
    \label{tab:tdv_ablations}
\end{table}

\section{Future Works and Broader Impact}
\label{sec:broader-impact} 
{\pa}'s weak inductive biases and joint frame-motion encoder design open up several new directions for future work.
First, deep learning approaches with weaker assumptions tend to scale more favorably with compute and data~\cite{sutton2019bitter, moutakanni2024you, chung2025shaping, dosovitskiy2020image}. Since {\pa} avoids augmentations, masking, contrastive objectives, and other strong inductive biases, it is positioned for stronger asymptotic performance than existing recipes.
Second, unlike existing visual representation learning approaches, {\pa} relies on no vision-specific techniques, such as masking or augmentation. Therefore, we believe {\pa} could be applied to any modality with high temporal consistency, including audio, proprioception, and touch. 
Third, {\pa} could enable efficient video encoding. Modern approaches for representing video typically pass every frame through a full image encoder. In contrast, using {\pa}, only the initial frame needs the frame encoder, with subsequent frames represented by composing the previous frame's representation with a lightweight motion encoder. This is similar to classical video codecs such as MPEG, which exploit temporal redundancy by storing keyframes and inter-frame deltas. 

\section{Limitations and Conclusion}
\label{sec:limitations}

\paragraph{Limitations.} While we achieved promising results with {\pa} in this work, there remain several limitations for further adoption.
First, while {\pa} matches existing approaches on dense spatial tasks, it does not achieve state-of-the-art results across the board. We view this as expected for a first attempt at representation learning without strong inductive biases, and anticipate that future work can build on the recipe to close the remaining gap.
Second, {\pa} did not achieve strong performance when measured on semantic benchmarks. We believe this is largely caused by a lack of inductive biases for learning invariances, such as local/global crops, which most existing visual representation learning approaches rely on~\cite{balestriero2023cookbook}.
Third, we found that scaling video data to larger video datasets than SomethingSomethingV2~\cite{goyal2017something} did not improve performance. We believe that this was caused by a lack of high-quality large scale open-source video data as well as our tuning of hyperparameters for performance on SomethingSomethingV2. We believe that with access to larger, higher-quality video datasets and better hyperparameters, {\pa} should scale further. Future work can search for more optimal hyperparameters that scale better to larger video datasets and model sizes. 

\paragraph{Conclusion.} In this work we proposed Temporal Difference in Vision (TDV), 
the first 
 approach for learning representations from videos without any supervision, raw pixel space reconstruction, or strong inductive biases. TDV achieves comparable or sometimes improved dense/spatial task performance to state-of-the-art visual representation learning recipes such as DINO~\cite{caron2021emerging} and iBOT~\cite{zhou2021ibot}, while not relying on strong inductive biases. As deep learning approaches leveraging weaker assumptions generally scale better, {\pa} lays the groundwork for potentially more scalable representation learning. 

\section{Acknowledgements}
We extend special thanks to Chris Hoang for his helpful discussions and advice; we are also thankful to Laude Institute for supporting this work.
This work is based upon work supported by the U.S. National Science Foundation Graduate Research Fellowship Program under Grant No. DGE 21-46756, U.S. DARPA ECOLE Program No. \#HR00112390060, DARPA ITM Program No. FA8650-23-C-7316, NSF Molecule Maker Lab Institute, an AI Institute for Molecular Discovery, Synthesis Strategy, and Manufacturing funded by the U.S. National Science Foundation under Awards No. 2019897 and 2505932, the AI Research Institutes program by National Science Foundation and the Institute of Education Sciences, U.S. Department of Education through Award No. 2229873 - AI Institute for Transforming Education for Children with Speech and Language Processing Challenges, and NSF NAIRR award. Any opinions, findings and conclusions or recommendations expressed in this material are those of the author(s) and do not necessarily reflect the views of the U.S. Government or the National Science Foundation. The U.S. Government is authorized to reproduce and distribute reprints for governmental purposes notwithstanding any copyright annotation therein. 
This research used the Delta and DeltaAI advanced computing and data resources, which are supported by the National Science Foundation (award OAC 2320345 and award OAC 2005572) and the State of Illinois. Delta and DeltaAI are joint efforts of the University of Illinois Urbana-Champaign and its National Center for Supercomputing Applications.

\clearpage
\newpage
{
    \small
    \bibliographystyle{unsrtnat}
    \bibliography{references}

@String(ECCV= {Eur. Conf. Comput. Vis.})

@String(AAAI = {AAAI})

@String(ECCV  = {ECCV})

@inproceedings{dosovitskiy2015flownet,
  title={FlowNet: Learning Optical Flow with Convolutional Networks},
  author={Dosovitskiy, Alexey and Fischer, Philipp and Ilg, Eddy and Hausser, Philip and Hazirbas, Caner and Golkov, Vladimir and Van Der Smagt, Patrick and Cremers, Daniel and Brox, Thomas},
  booktitle={Proceedings of the IEEE International Conference on Computer Vision},
  pages={2758--2766},
  year={2015}
}

@inproceedings{grauman2022ego4d,
  title={Ego4D: Around the World in 3,000 Hours of Egocentric Video},
  author={Grauman, Kristen and Westbury, Andrew and Byrne, Eugene and Chavis, Zachary and Furnari, Antonino and Girdhar, Rohit and Hamburger, Jackson and Jiang, Hao and Liu, Miao and Liu, Xingyu and others},
  booktitle={Proceedings of the IEEE/CVF Conference on Computer Vision and Pattern Recognition},
  pages={18995--19012},
  year={2022}
}

@inproceedings{perez2018film,
  title={{FiLM}: Visual Reasoning with a General Conditioning Layer},
  author={Perez, Ethan and Strub, Florian and De Vries, Harm and Dumoulin, Vincent and Courville, Aaron},
  booktitle={Proceedings of the AAAI Conference on Artificial Intelligence},
  volume={32},
  year={2018}
}

@misc{mmseg2020,
  title={{MMSegmentation}: OpenMMLab Semantic Segmentation Toolbox and Benchmark},
  author={{MMSegmentation Contributors}},
  howpublished={\url{https://github.com/open-mmlab/mmsegmentation}},
  year={2020}
}

@inproceedings{zhou2017scene,
  title={Scene Parsing through ADE20K Dataset},
  author={Zhou, Bolei and Zhao, Hang and Puig, Xavier and Fidler, Sanja and Barriuso, Adela and Torralba, Antonio},
  booktitle={Proceedings of the IEEE Conference on Computer Vision and Pattern Recognition},
  pages={633--641},
  year={2017}
}

@inproceedings{cordts2016cityscapes,
  title={The Cityscapes Dataset for Semantic Urban Scene Understanding},
  author={Cordts, Marius and Omran, Mohamed and Ramos, Sebastian and Rehfeld, Timo and Enzweiler, Markus and Benenson, Rodrigo and Franke, Uwe and Roth, Stefan and Schiele, Bernt},
  booktitle={Proceedings of the IEEE Conference on Computer Vision and Pattern Recognition},
  pages={3213--3223},
  year={2016}
}

@misc{Farre2024FineVideo,
  title={FineVideo},
  author={Farré, Miquel and Marafioti, Andi and Tunstall, Lewis and Von Werra, Leandro and Wolf, Thomas},
  year={2024},
  howpublished={\url{https://huggingface.co/datasets/HuggingFaceFV/finevideo}},
}

@article{felleman1991distributed,
  title={Distributed hierarchical processing in the primate cerebral cortex.},
  author={Felleman, Daniel J and Van Essen, David C},
  journal={Cerebral cortex (New York, NY: 1991)},
  volume={1},
  number={1},
  pages={1--47},
  year={1991}
}

@article{goodale1992separate,
  title={Separate visual pathways for perception and action},
  author={Goodale, Melvyn A and Milner, A David},
  journal={Trends in neurosciences},
  volume={15},
  number={1},
  pages={20--25},
  year={1992},
  publisher={Elsevier}
}

@article{russakovsky2015imagenet,
  title={Imagenet large scale visual recognition challenge},
  author={Russakovsky, Olga and Deng, Jia and Su, Hao and Krause, Jonathan and Satheesh, Sanjeev and Ma, Sean and Huang, Zhiheng and Karpathy, Andrej and Khosla, Aditya and Bernstein, Michael and others},
  journal={International journal of computer vision},
  volume={115},
  pages={211--252},
  year={2015},
  publisher={Springer}
}

@inproceedings{rombach2022high,
  title={High-resolution image synthesis with latent diffusion models},
  author={Rombach, Robin and Blattmann, Andreas and Lorenz, Dominik and Esser, Patrick and Ommer, Bj{\"o}rn},
  booktitle={Proceedings of the IEEE/CVF conference on computer vision and pattern recognition},
  pages={10684--10695},
  year={2022}
}

@article{su2024roformer,
  title={Roformer: Enhanced transformer with rotary position embedding},
  author={Su, Jianlin and Ahmed, Murtadha and Lu, Yu and Pan, Shengfeng and Bo, Wen and Liu, Yunfeng},
  journal={Neurocomputing},
  volume={568},
  pages={127063},
  year={2024},
  publisher={Elsevier}
}

@article{guo2025deepseek,
  title={Deepseek-r1: Incentivizing reasoning capability in llms via reinforcement learning},
  author={Guo, Daya and Yang, Dejian and Zhang, Haowei and Song, Junxiao and Zhang, Ruoyu and Xu, Runxin and Zhu, Qihao and Ma, Shirong and Wang, Peiyi and Bi, Xiao and others},
  journal={arXiv preprint arXiv:2501.12948},
  year={2025}
}

@article{grill2020bootstrap,
  title={Bootstrap your own latent-a new approach to self-supervised learning},
  author={Grill, Jean-Bastien and Strub, Florian and Altch{\'e}, Florent and Tallec, Corentin and Richemond, Pierre and Buchatskaya, Elena and Doersch, Carl and Avila Pires, Bernardo and Guo, Zhaohan and Gheshlaghi Azar, Mohammad and others},
  journal={Advances in neural information processing systems},
  volume={33},
  pages={21271--21284},
  year={2020}
}

@misc{peebles2023scalable,
      title={Scalable Diffusion Models with Transformers}, 
      author={William Peebles and Saining Xie},
      year={2023},
      eprint={2212.09748},
      archivePrefix={arXiv},
      primaryClass={cs.CV}
}

@misc{chen2020simple,
      title={A Simple Framework for Contrastive Learning of Visual Representations}, 
      author={Ting Chen and Simon Kornblith and Mohammad Norouzi and Geoffrey Hinton},
      year={2020},
      eprint={2002.05709},
      archivePrefix={arXiv},
      primaryClass={cs.LG}
}

@article{kay2017kinetics,
  title={The kinetics human action video dataset},
  author={Kay, Will and Carreira, Joao and Simonyan, Karen and Zhang, Brian and Hillier, Chloe and Vijayanarasimhan, Sudheendra and Viola, Fabio and Green, Tim and Back, Trevor and Natsev, Paul and others},
  journal={arXiv preprint arXiv:1705.06950},
  year={2017}
}

@inproceedings{he2020momentum,
  title={Momentum contrast for unsupervised visual representation learning},
  author={He, Kaiming and Fan, Haoqi and Wu, Yuxin and Xie, Saining and Girshick, Ross},
  booktitle={Proceedings of the IEEE/CVF conference on computer vision and pattern recognition},
  pages={9729--9738},
  year={2020}
}

@misc{oquab2023dinov2,
      title={DINOv2: Learning Robust Visual Features without Supervision}, 
      author={Maxime Oquab and Timothée Darcet and Théo Moutakanni and Huy Vo and Marc Szafraniec and Vasil Khalidov and Pierre Fernandez and Daniel Haziza and Francisco Massa and Alaaeldin El-Nouby and Mahmoud Assran and Nicolas Ballas and Wojciech Galuba and Russell Howes and Po-Yao Huang and Shang-Wen Li and Ishan Misra and Michael Rabbat and Vasu Sharma and Gabriel Synnaeve and Hu Xu and Hervé Jegou and Julien Mairal and Patrick Labatut and Armand Joulin and Piotr Bojanowski},
      year={2023},
      eprint={2304.07193},
      archivePrefix={arXiv},
      primaryClass={cs.CV}
}

@misc{assran2023selfsupervised,
      title={Self-Supervised Learning from Images with a Joint-Embedding Predictive Architecture}, 
      author={Mahmoud Assran and Quentin Duval and Ishan Misra and Piotr Bojanowski and Pascal Vincent and Michael Rabbat and Yann LeCun and Nicolas Ballas},
      year={2023},
      eprint={2301.08243},
      archivePrefix={arXiv},
      primaryClass={cs.CV}
}

@misc{bardes2022vicreg,
      title={VICReg: Variance-Invariance-Covariance Regularization for Self-Supervised Learning}, 
      author={Adrien Bardes and Jean Ponce and Yann LeCun},
      year={2022},
      eprint={2105.04906},
      archivePrefix={arXiv},
      primaryClass={cs.CV}
}

@misc{he2021masked,
      title={Masked Autoencoders Are Scalable Vision Learners}, 
      author={Kaiming He and Xinlei Chen and Saining Xie and Yanghao Li and Piotr Dollár and Ross Girshick},
      year={2021},
      eprint={2111.06377},
      archivePrefix={arXiv},
      primaryClass={cs.CV}
}

@misc{balestriero2023cookbook,
      title={A Cookbook of Self-Supervised Learning}, 
      author={Randall Balestriero and Mark Ibrahim and Vlad Sobal and Ari Morcos and Shashank Shekhar and Tom Goldstein and Florian Bordes and Adrien Bardes and Gregoire Mialon and Yuandong Tian and Avi Schwarzschild and Andrew Gordon Wilson and Jonas Geiping and Quentin Garrido and Pierre Fernandez and Amir Bar and Hamed Pirsiavash and Yann LeCun and Micah Goldblum},
      year={2023},
      eprint={2304.12210},
      archivePrefix={arXiv},
      primaryClass={cs.LG}
}

@misc{caron2021emerging,
      title={Emerging Properties in Self-Supervised Vision Transformers}, 
      author={Mathilde Caron and Hugo Touvron and Ishan Misra and Hervé Jégou and Julien Mairal and Piotr Bojanowski and Armand Joulin},
      year={2021},
      eprint={2104.14294},
      archivePrefix={arXiv},
      primaryClass={cs.CV}
}

@inproceedings{goyal2017something,
  title={The" something something" video database for learning and evaluating visual common sense},
  author={Goyal, Raghav and Ebrahimi Kahou, Samira and Michalski, Vincent and Materzynska, Joanna and Westphal, Susanne and Kim, Heuna and Haenel, Valentin and Fruend, Ingo and Yianilos, Peter and Mueller-Freitag, Moritz and others},
  booktitle={Proceedings of the IEEE international conference on computer vision},
  pages={5842--5850},
  year={2017}
}

@article{hoffmann2022training,
  title={Training compute-optimal large language models},
  author={Hoffmann, Jordan and Borgeaud, Sebastian and Mensch, Arthur and Buchatskaya, Elena and Cai, Trevor and Rutherford, Eliza and Casas, Diego de Las and Hendricks, Lisa Anne and Welbl, Johannes and Clark, Aidan and others},
  journal={arXiv preprint arXiv:2203.15556},
  year={2022}
}

@article{kaplan2020scaling,
  title={Scaling laws for neural language models},
  author={Kaplan, Jared and McCandlish, Sam and Henighan, Tom and Brown, Tom B and Chess, Benjamin and Child, Rewon and Gray, Scott and Radford, Alec and Wu, Jeffrey and Amodei, Dario},
  journal={arXiv preprint arXiv:2001.08361},
  year={2020}
}

@article{bardes2024revisiting,
  title={Revisiting feature prediction for learning visual representations from video},
  author={Bardes, Adrien and Garrido, Quentin and Ponce, Jean and Chen, Xinlei and Rabbat, Michael and LeCun, Yann and Assran, Mahmoud and Ballas, Nicolas},
  journal={arXiv preprint arXiv:2404.08471},
  year={2024}
}

@article{lecun2022path,
  title={A path towards autonomous machine intelligence version 0.9. 2, 2022-06-27},
  author={LeCun, Yann},
  journal={Open Review},
  volume={62},
  year={2022}
}

@misc{openai2023gpt4,
      title={GPT-4 Technical Report}, 
      author={OpenAI},
      year={2023},
      eprint={2303.08774},
      archivePrefix={arXiv},
      primaryClass={cs.CL}
}

@misc{touvron2023llama,
      title={Llama 2: Open Foundation and Fine-Tuned Chat Models}, 
      author={Hugo Touvron and Louis Martin and Kevin Stone and Peter Albert and Amjad Almahairi and Yasmine Babaei and Nikolay Bashlykov and Soumya Batra and Prajjwal Bhargava and Shruti Bhosale and Dan Bikel and Lukas Blecher and Cristian Canton Ferrer and Moya Chen and Guillem Cucurull and David Esiobu and Jude Fernandes and Jeremy Fu and Wenyin Fu and Brian Fuller and Cynthia Gao and Vedanuj Goswami and Naman Goyal and Anthony Hartshorn and Saghar Hosseini and Rui Hou and Hakan Inan and Marcin Kardas and Viktor Kerkez and Madian Khabsa and Isabel Kloumann and Artem Korenev and Punit Singh Koura and Marie-Anne Lachaux and Thibaut Lavril and Jenya Lee and Diana Liskovich and Yinghai Lu and Yuning Mao and Xavier Martinet and Todor Mihaylov and Pushkar Mishra and Igor Molybog and Yixin Nie and Andrew Poulton and Jeremy Reizenstein and Rashi Rungta and Kalyan Saladi and Alan Schelten and Ruan Silva and Eric Michael Smith and Ranjan Subramanian and Xiaoqing Ellen Tan and Binh Tang and Ross Taylor and Adina Williams and Jian Xiang Kuan and Puxin Xu and Zheng Yan and Iliyan Zarov and Yuchen Zhang and Angela Fan and Melanie Kambadur and Sharan Narang and Aurelien Rodriguez and Robert Stojnic and Sergey Edunov and Thomas Scialom},
      year={2023},
      eprint={2307.09288},
      archivePrefix={arXiv},
      primaryClass={cs.CL}
}

@article{bardes2023v,
  title={V-JEPA: Latent Video Prediction for Visual Representation Learning},
  author={Bardes, Adrien and Garrido, Quentin and Ponce, Jean and Chen, Xinlei and Rabbat, Michael and LeCun, Yann and Assran, Mido and Ballas, Nicolas},
  year={2023}
}

@article{ha2018world,
  title={World models},
  author={Ha, David and Schmidhuber, J{\"u}rgen},
  journal={arXiv preprint arXiv:1803.10122},
  year={2018}
}

@article{oord2018representation,
  title={Representation learning with contrastive predictive coding},
  author={Oord, Aaron van den and Li, Yazhe and Vinyals, Oriol},
  journal={arXiv preprint arXiv:1807.03748},
  year={2018}
}

@article{gladstone2025energy,
  title={Energy-Based Transformers are Scalable Learners and Thinkers},
  author={Gladstone, Alexi and Nanduru, Ganesh and Islam, Md Mofijul and Han, Peixuan and Ha, Hyeonjeong and Chadha, Aman and Du, Yilun and Ji, Heng and Li, Jundong and Iqbal, Tariq},
  journal={arXiv preprint arXiv:2507.02092},
  year={2025}
}

@inproceedings{radford2021learning,
  title={Learning transferable visual models from natural language supervision},
  author={Radford, Alec and Kim, Jong Wook and Hallacy, Chris and Ramesh, Aditya and Goh, Gabriel and Agarwal, Sandhini and Sastry, Girish and Askell, Amanda and Mishkin, Pamela and Clark, Jack and others},
  booktitle={International conference on machine learning},
  pages={8748--8763},
  year={2021},
  organization={PmLR}
}

@article{sutton2019bitter,
  title={The bitter lesson},
  author={Sutton, Richard},
  journal={Incomplete Ideas (blog)},
  volume={13},
  number={1},
  pages={38},
  year={2019}
}

@article{redmon2018yolov3,
  title={Yolov3: An incremental improvement},
  author={Redmon, Joseph and Farhadi, Ali},
  journal={arXiv preprint arXiv:1804.02767},
  year={2018}
}

@article{krizhevsky2012imagenet,
  title={Imagenet classification with deep convolutional neural networks},
  author={Krizhevsky, Alex and Sutskever, Ilya and Hinton, Geoffrey E},
  journal={Advances in neural information processing systems},
  volume={25},
  year={2012}
}

@article{simeoni2025dinov3,
  title={Dinov3},
  author={Sim{\'e}oni, Oriane and Vo, Huy V and Seitzer, Maximilian and Baldassarre, Federico and Oquab, Maxime and Jose, Cijo and Khalidov, Vasil and Szafraniec, Marc and Yi, Seungeun and Ramamonjisoa, Micha{\"e}l and others},
  journal={arXiv preprint arXiv:2508.10104},
  year={2025}
}

@inproceedings{he2016deep,
  title={Deep residual learning for image recognition},
  author={He, Kaiming and Zhang, Xiangyu and Ren, Shaoqing and Sun, Jian},
  booktitle={Proceedings of the IEEE conference on computer vision and pattern recognition},
  pages={770--778},
  year={2016}
}

@article{dosovitskiy2020image,
  title={An image is worth 16x16 words: Transformers for image recognition at scale},
  author={Dosovitskiy, Alexey},
  journal={arXiv preprint arXiv:2010.11929},
  year={2020}
}

@article{reader2002social,
  title={Social intelligence, innovation, and enhanced brain size in primates},
  author={Reader, Simon M and Laland, Kevin N},
  journal={Proceedings of the National Academy of Sciences},
  volume={99},
  number={7},
  pages={4436--4441},
  year={2002},
  publisher={The National Academy of Sciences}
}

@article{gomez2024evolution,
  title={The evolution of human altriciality and brain development in comparative context},
  author={G{\'o}mez-Robles, Aida and Nicolaou, Christos and Smaers, Jeroen B and Sherwood, Chet C},
  journal={Nature Ecology \& Evolution},
  volume={8},
  number={1},
  pages={133--146},
  year={2024},
  publisher={Nature Publishing Group UK London}
}

@book{tinbergen2020study,
  title={The study of instinct},
  author={Tinbergen, Nikolaas},
  year={2020},
  publisher={Pygmalion Press, an imprint of Plunkett Lake Press}
}

@article{balestriero2025lejepa,
  title={Lejepa: Provable and scalable self-supervised learning without the heuristics},
  author={Balestriero, Randall and LeCun, Yann},
  journal={arXiv preprint arXiv:2511.08544},
  year={2025}
}

@inproceedings{noroozi2016unsupervised,
  title={Unsupervised learning of visual representations by solving jigsaw puzzles},
  author={Noroozi, Mehdi and Favaro, Paolo},
  booktitle={European conference on computer vision},
  pages={69--84},
  year={2016},
  organization={Springer}
}

@article{gidaris2018unsupervised,
  title={Unsupervised representation learning by predicting image rotations},
  author={Gidaris, Spyros and Singh, Praveer and Komodakis, Nikos},
  journal={arXiv preprint arXiv:1803.07728},
  year={2018}
}

@inproceedings{zhang2016colorful,
  title={Colorful image colorization},
  author={Zhang, Richard and Isola, Phillip and Efros, Alexei A},
  booktitle={European conference on computer vision},
  pages={649--666},
  year={2016},
  organization={Springer}
}

@article{moutakanni2024you,
  title={You don’t need domain-specific data augmentations when scaling self-supervised learning},
  author={Moutakanni, Th{\'e}o and Oquab, Maxime and Szafraniec, Marc and Vakalopoulou, Maria and Bojanowski, Piotr},
  journal={Advances in Neural Information Processing Systems},
  volume={37},
  pages={116106--116125},
  year={2024}
}

@article{caron2020unsupervised,
  title={Unsupervised learning of visual features by contrasting cluster assignments},
  author={Caron, Mathilde and Misra, Ishan and Mairal, Julien and Goyal, Priya and Bojanowski, Piotr and Joulin, Armand},
  journal={Advances in neural information processing systems},
  volume={33},
  pages={9912--9924},
  year={2020}
}

@article{carreira2024scaling,
  title={Scaling 4d representations},
  author={Carreira, Jo{\~a}o and Gokay, Dilara and King, Michael and Zhang, Chuhan and Rocco, Ignacio and Mahendran, Aravindh and Keck, Thomas Albert and Heyward, Joseph and Koppula, Skanda and Pot, Etienne and others},
  journal={arXiv preprint arXiv:2412.15212},
  year={2024}
}

@article{darcet2025cluster,
  title={Cluster and predict latent patches for improved masked image modeling},
  author={Darcet, Timoth{\'e}e and Baldassarre, Federico and Oquab, Maxime and Mairal, Julien and Bojanowski, Piotr},
  journal={arXiv preprint arXiv:2502.08769},
  year={2025}
}

@article{hoang2025midway,
  title={Midway Network: Learning Representations for Recognition and Motion from Latent Dynamics},
  author={Hoang, Christopher and Ren, Mengye},
  journal={arXiv preprint arXiv:2510.05558},
  year={2025}
}

@inproceedings{feng2023mutual,
  title={Mutual information-based temporal difference learning for human pose estimation in video},
  author={Feng, Runyang and Gao, Yixing and Ma, Xueqing and Tse, Tze Ho Elden and Chang, Hyung Jin},
  booktitle={Proceedings of the IEEE/CVF Conference on Computer Vision and Pattern Recognition},
  pages={17131--17141},
  year={2023}
}

@article{simonyan2014two,
  title={Two-stream convolutional networks for action recognition in videos},
  author={Simonyan, Karen and Zisserman, Andrew},
  journal={Advances in neural information processing systems},
  volume={27},
  year={2014}
}

@inproceedings{recasens2021broaden,
  title={Broaden your views for self-supervised video learning},
  author={Recasens, Adria and Luc, Pauline and Alayrac, Jean-Baptiste and Wang, Luyu and Strub, Florian and Tallec, Corentin and Malinowski, Mateusz and P{\u{a}}tr{\u{a}}ucean, Viorica and Altch{\'e}, Florent and Valko, Michal and others},
  booktitle={Proceedings of the IEEE/CVF international conference on computer vision},
  pages={1255--1265},
  year={2021}
}

@article{bardes2023mc,
  title={Mc-jepa: A joint-embedding predictive architecture for self-supervised learning of motion and content features},
  author={Bardes, Adrien and Ponce, Jean and LeCun, Yann},
  journal={arXiv preprint arXiv:2307.12698},
  year={2023}
}

@inproceedings{wang2021tdn,
  title={Tdn: Temporal difference networks for efficient action recognition},
  author={Wang, Limin and Tong, Zhan and Ji, Bin and Wu, Gangshan},
  booktitle={Proceedings of the IEEE/CVF conference on computer vision and pattern recognition},
  pages={1895--1904},
  year={2021}
}

@article{balestriero2024learning,
  title={Learning by reconstruction produces uninformative features for perception},
  author={Balestriero, Randall and LeCun, Yann},
  journal={arXiv preprint arXiv:2402.11337},
  year={2024}
}

@article{littwin2024jepa,
  title={How jepa avoids noisy features: The implicit bias of deep linear self distillation networks},
  author={Littwin, Etai and Saremi, Omid and Advani, Madhu and Thilak, Vimal and Nakkiran, Preetum and Huang, Chen and Susskind, Joshua},
  journal={Advances in Neural Information Processing Systems},
  volume={37},
  pages={91300--91336},
  year={2024}
}

@article{assran2025v,
  title={V-jepa 2: Self-supervised video models enable understanding, prediction and planning},
  author={Assran, Mido and Bardes, Adrien and Fan, David and Garrido, Quentin and Howes, Russell and Muckley, Matthew and Rizvi, Ammar and Roberts, Claire and Sinha, Koustuv and Zholus, Artem and others},
  journal={arXiv preprint arXiv:2506.09985},
  year={2025}
}

@article{rolfe2013discriminative,
  title={Discriminative recurrent sparse auto-encoders},
  author={Rolfe, Jason Tyler and LeCun, Yann},
  journal={arXiv preprint arXiv:1301.3775},
  year={2013}
}

@inproceedings{vincent2008extracting,
  title={Extracting and composing robust features with denoising autoencoders},
  author={Vincent, Pascal and Larochelle, Hugo and Bengio, Yoshua and Manzagol, Pierre-Antoine},
  booktitle={Proceedings of the 25th international conference on Machine learning},
  pages={1096--1103},
  year={2008}
}

@article{zhou2021ibot,
  title={ibot: Image bert pre-training with online tokenizer},
  author={Zhou, Jinghao and Wei, Chen and Wang, Huiyu and Shen, Wei and Xie, Cihang and Yuille, Alan and Kong, Tao},
  journal={arXiv preprint arXiv:2111.07832},
  year={2021}
}

@inproceedings{weinzaepfel2023croco,
  title={Croco v2: Improved cross-view completion pre-training for stereo matching and optical flow},
  author={Weinzaepfel, Philippe and Lucas, Thomas and Leroy, Vincent and Cabon, Yohann and Arora, Vaibhav and Br{\'e}gier, Romain and Csurka, Gabriela and Antsfeld, Leonid and Chidlovskii, Boris and Revaud, J{\'e}r{\^o}me},
  booktitle={Proceedings of the IEEE/CVF International Conference on Computer Vision},
  pages={17969--17980},
  year={2023}
}

@inproceedings{zbontar2021barlow,
  title={Barlow twins: Self-supervised learning via redundancy reduction},
  author={Zbontar, Jure and Jing, Li and Misra, Ishan and LeCun, Yann and Deny, St{\'e}phane},
  booktitle={International conference on machine learning},
  pages={12310--12320},
  year={2021},
  organization={PMLR}
}

@inproceedings{van2016pixel,
  title={Pixel recurrent neural networks},
  author={Van Den Oord, A{\"a}ron and Kalchbrenner, Nal and Kavukcuoglu, Koray},
  booktitle={International conference on machine learning},
  pages={1747--1756},
  year={2016},
  organization={PMLR}
}

@misc{chung2025shaping,
  author       = {Chung, Hyung Won},
  title        = {Shaping the Future of {AI} from the History of {Transformer}},
  year         = {2025},
  howpublished = {\url{https://www.youtube.com/watch?v=orDKvo8h71o}},
  note         = {Talk}
}

@inproceedings{butler2012naturalistic,
  title={A naturalistic open source movie for optical flow evaluation},
  author={Butler, Daniel J and Wulff, Jonas and Stanley, Garrett B and Black, Michael J},
  booktitle={European conference on computer vision},
  pages={611--625},
  year={2012},
  organization={Springer}
}

@inproceedings{mayer2016large,
  title={A large dataset to train convolutional networks for disparity, optical flow, and scene flow estimation},
  author={Mayer, Nikolaus and Ilg, Eddy and Hausser, Philip and Fischer, Philipp and Cremers, Daniel and Dosovitskiy, Alexey and Brox, Thomas},
  booktitle={Proceedings of the IEEE conference on computer vision and pattern recognition},
  pages={4040--4048},
  year={2016}
}

@inproceedings{xiao2018unified,
  title={Unified perceptual parsing for scene understanding},
  author={Xiao, Tete and Liu, Yingcheng and Zhou, Bolei and Jiang, Yuning and Sun, Jian},
  booktitle={Proceedings of the European conference on computer vision (ECCV)},
  pages={418--434},
  year={2018}
}

@article{dukas2008evolutionary,
  title={Evolutionary biology of insect learning},
  author={Dukas, Reuven},
  journal={Annu. Rev. Entomol.},
  volume={53},
  number={1},
  pages={145--160},
  year={2008},
  publisher={Annual Reviews}
}

@article{einstein1905electrodynamics,
  title={On the electrodynamics of moving bodies},
  author={Einstein, Albert and others},
  journal={Annalen der physik},
  volume={17},
  number={10},
  pages={891--921},
  year={1905}
}

@article{d2018causality,
  title={Causality re-established},
  author={D’Ariano, Giacomo Mauro},
  journal={Philosophical Transactions of the Royal Society A: Mathematical, Physical and Engineering Sciences},
  volume={376},
  number={2123},
  year={2018},
  publisher={The Royal Society}
}

@article{bell1964einstein,
  title={On the einstein podolsky rosen paradox},
  author={Bell, John S},
  journal={Physics Physique Fizika},
  volume={1},
  number={3},
  pages={195},
  year={1964},
  publisher={APS}
}

@article{sutton1988learning,
  title={Learning to predict by the methods of temporal differences},
  author={Sutton, Richard S},
  journal={Machine learning},
  volume={3},
  number={1},
  pages={9--44},
  year={1988},
  publisher={Springer}
}

@article{maes2026leworldmodel,
  title={Leworldmodel: Stable end-to-end joint-embedding predictive architecture from pixels},
  author={Maes, Lucas and Lidec, Quentin Le and Scieur, Damien and LeCun, Yann and Balestriero, Randall},
  journal={arXiv preprint arXiv:2603.19312},
  year={2026}
}

@article{ericsson2021self,
  title={Why do self-supervised models transfer? investigating the impact of invariance on downstream tasks},
  author={Ericsson, Linus and Gouk, Henry and Hospedales, Timothy M},
  journal={arXiv preprint arXiv:2111.11398},
  year={2021}
}

@article{xiao2020should,
  title={What should not be contrastive in contrastive learning},
  author={Xiao, Tete and Wang, Xiaolong and Efros, Alexei A and Darrell, Trevor},
  journal={arXiv preprint arXiv:2008.05659},
  year={2020}
}

@inproceedings{misra2020self,
  title={Self-supervised learning of pretext-invariant representations},
  author={Misra, Ishan and Maaten, Laurens van der},
  booktitle={Proceedings of the IEEE/CVF conference on computer vision and pattern recognition},
  pages={6707--6717},
  year={2020}
}

@article{purushwalkam2020demystifying,
  title={Demystifying contrastive self-supervised learning: Invariances, augmentations and dataset biases},
  author={Purushwalkam, Senthil and Gupta, Abhinav},
  journal={Advances in Neural Information Processing Systems},
  volume={33},
  pages={3407--3418},
  year={2020}
}

@article{von2021self,
  title={Self-supervised learning with data augmentations provably isolates content from style},
  author={Von K{\"u}gelgen, Julius and Sharma, Yash and Gresele, Luigi and Brendel, Wieland and Sch{\"o}lkopf, Bernhard and Besserve, Michel and Locatello, Francesco},
  journal={Advances in neural information processing systems},
  volume={34},
  pages={16451--16467},
  year={2021}
}

@article{wiegand2003overview,
  title={Overview of the H. 264/AVC video coding standard},
  author={Wiegand, Thomas and Sullivan, Gary J and Bjontegaard, Gisle and Luthra, Ajay},
  journal={IEEE Transactions on circuits and systems for video technology},
  volume={13},
  number={7},
  pages={560--576},
  year={2003},
  publisher={IEEE}
}

@inproceedings{han2019video,
  title={Video representation learning by dense predictive coding},
  author={Han, Tengda and Xie, Weidi and Zisserman, Andrew},
  booktitle={Proceedings of the IEEE/CVF international conference on computer vision workshops},
  pages={0--0},
  year={2019}
}

@inproceedings{bruce2024genie,
  title={Genie: Generative interactive environments},
  author={Bruce, Jake and Dennis, Michael D and Edwards, Ashley and Parker-Holder, Jack and Shi, Yuge and Hughes, Edward and Lai, Matthew and Mavalankar, Aditi and Steigerwald, Richie and Apps, Chris and others},
  booktitle={Forty-first International Conference on Machine Learning},
  year={2024}
}

@article{rao1999predictive,
  title={Predictive coding in the visual cortex: a functional interpretation of some extra-classical receptive-field effects},
  author={Rao, Rajesh PN and Ballard, Dana H},
  journal={Nature neuroscience},
  volume={2},
  number={1},
  pages={79--87},
  year={1999},
  publisher={Nature Publishing Group}
}

@article{friston2010free,
  title={The free-energy principle: a unified brain theory?},
  author={Friston, Karl},
  journal={Nature reviews neuroscience},
  volume={11},
  number={2},
  pages={127--138},
  year={2010},
  publisher={Nature publishing group}
}

@inproceedings{Ungerleider1982TwoCV,
  title={Two cortical visual systems},
  author={Leslie G. Ungerleider},
  year={1982},
  url={https://api.semanticscholar.org/CorpusID:142774685}
}

@article{bastos2012canonical,
  title={Canonical microcircuits for predictive coding},
  author={Bastos, Andre M and Usrey, W Martin and Adams, Rick A and Mangun, George R and Fries, Pascal and Friston, Karl J},
  journal={Neuron},
  volume={76},
  number={4},
  pages={695--711},
  year={2012},
  publisher={Elsevier}
}

@article{henaff2021primary,
  title={Primary visual cortex straightens natural video trajectories},
  author={H{\'e}naff, Olivier J and Bai, Yoon and Charlton, Julie A and Nauhaus, Ian and Simoncelli, Eero P and Goris, Robbe LT},
  journal={Nature communications},
  volume={12},
  number={1},
  pages={5982},
  year={2021},
  publisher={Nature Publishing Group UK London}
}

@article{henaff2019perceptual,
  title={Perceptual straightening of natural videos},
  author={H{\'e}naff, Olivier J and Goris, Robbe LT and Simoncelli, Eero P},
  journal={Nature neuroscience},
  volume={22},
  number={6},
  pages={984--991},
  year={2019},
  publisher={Nature Publishing Group US New York}
}
}

\clearpage
\newpage
\appendix

\renewcommand{\thesection}{\Alph{section}}          
\renewcommand{\thefigure}{\Alph{section}\arabic{figure}} 
\renewcommand{\thetable}{\Alph{section}\arabic{table}}   

\setcounter{section}{0}
\counterwithin{figure}{section}   
\counterwithin{table}{section}

\clearpage
\section{Additional Intuition}
\label{sec:additional_intuition}

To our knowledge, TDV is the first purely unsupervised visual representation learning approach that avoids \emph{all} of the following inductive biases simultaneously: raw data reconstruction~\cite{he2021masked}, aligned data with other modalities~\cite{radford2021learning}, hand-crafted pretext tasks \cite{noroozi2016unsupervised, gidaris2018unsupervised, zhang2016colorful}, contrastive learning~\cite{chen2020simple}, clustering objectives~\cite{caron2020unsupervised, darcet2025cluster}, augmentations~\cite{chen2020simple, caron2021emerging}, explicit invariances~\cite{bardes2022vicreg}, redundancy reduction (i.e., Variance or Covariance regularization~\cite{bardes2022vicreg}), and cropping or masking~\cite{oquab2023dinov2, balestriero2025lejepa}.
While each of these techniques has driven significant progress in representation learning, each also introduces assumptions that can become limiting as data and compute scale. We discuss the limitations of each below; prior work has raised similar concerns regarding how invariances and pretext biases can affect performance~\cite{ericsson2021self, xiao2020should}.

\paragraph{Image Augmentations.} Augmentation-based approaches~\cite{chen2020simple, caron2021emerging, oquab2023dinov2} often pull together differently augmented views of the same image, implicitly treating whatever the augmentation changed as irrelevant to semantics. However, which factors are irrelevant depends on the downstream task. Color jitter encourages invariance to color, which can help coarse classification but is unhelpful for tasks where color is informative, such as bird species, ripeness, or material recognition. Similarly, spatial augmentations encourage invariance to location, which is precisely the information that detection must preserve. Any forced invariance therefore tends to help some tasks while hurting others, and we expect this trade-off to become more pronounced at the limit of learning~\cite{ericsson2021self, xiao2020should}.

\paragraph{Masking.} Masking-based approaches~\cite{oquab2023dinov2, balestriero2025lejepa, zhou2021ibot} face a closely related issue: they encourage an image and a heavily masked version of it to map to nearly the same representation, even though much of the visual content has been removed. This can collapse semantically distinct inputs together and reduce spatial awareness. Detection illustrates the concern: two images of the same object at different positions should ideally produce different representations so that location can be recovered, whereas masking pushes them toward a shared representation. Section~\ref{sec:motivating_weaker_assumptions} provides an empirical version of this argument, showing that the optimal masking ratio decreases as data scale grows.

\paragraph{Contrastive Learning.} Contrastive approaches~\cite{chen2020simple, he2020momentum} push randomly sampled images apart in representation space. This is only approximately valid, as two random images from a natural distribution often share content (e.g., two outdoor scenes, two dogs, or two faces), and separating them can discard useful structure. Contrastive learning also relies on large numbers of negative samples, which makes scaling costly in both batch size and memory.

\paragraph{Raw Pixel Reconstruction.} Raw reconstruction-based approaches~\cite{he2021masked, rolfe2013discriminative, vincent2008extracting} require the representation to retain enough information to rebuild every pixel, including texture, lighting, and background detail that may be unrelated to the scene content. Both theory~\cite{balestriero2024learning} and experiments~\cite{littwin2024jepa} indicate that pixel-space targets yield features that are less useful for perception than latent-space targets. Requiring the representation to encode all such detail can also limit how abstract it becomes, since fine-grained appearance must be retained somewhere.

\paragraph{Cross-Modal Alignment.} Approaches such as CLIP~\cite{radford2021learning} learn vision representations by aligning them with text. This works well when paired data is plentiful, but it ties the visual representation to whatever the accompanying text describes. Captions tend to emphasize objects and actions while omitting texture, geometry, lighting, and spatial layout, so the resulting features can be biased toward nameable content and weaker on dense spatial properties that text rarely describes. The approach also requires paired modalities, which is itself a strong data assumption.

\paragraph{Hand-Crafted Pretext Tasks.} Pretext objectives such as jigsaw puzzles~\cite{noroozi2016unsupervised}, rotation prediction~\cite{gidaris2018unsupervised}, and colorization~\cite{zhang2016colorful} solve a synthetic problem in the hope that the features learned along the way transfer. A limitation is that the model only needs to learn whatever is sufficient for that specific task: rotation prediction, for instance, can rely on a few orientation cues such as sky position or text, leaving the rest of the representation underdeveloped. More broadly, the designer must anticipate what makes a useful pretext, and each choice reflects a particular view of which visual structure matters.

\paragraph{Clustering Objectives.} Clustering-based approaches~\cite{caron2020unsupervised, darcet2025cluster} assign images to discrete prototypes and train the encoder to be consistent with those assignments. This assumes that the data falls into a fixed number of well-separated clusters, which is only approximately true for natural images: semantic categories overlap, vary continuously, and exist at multiple granularities simultaneously. The number of clusters becomes a hyperparameter that bounds the granularity the representation can reach, and clustering pipelines often require additional machinery (e.g., Sinkhorn balancing or queue tricks) to avoid degenerate solutions.

\paragraph{Explicit Invariances and Redundancy Reduction.} Methods such as VICReg~\cite{bardes2022vicreg} avoid contrastive negatives by directly regularizing feature statistics across a batch: an invariance term between augmented views, together with variance and covariance terms computed over batch dimensions to prevent collapse. The invariance term shares the limitation of augmentation-based methods, as features are encouraged to ignore whatever the augmentation changed. The variance and covariance terms introduce a separate consideration: they assume that a batch of independently sampled images should produce features that are well-spread and decorrelated, which holds only when batches are large and diverse enough to approximate the data distribution. Small batches, batches with correlated content (e.g., consecutive video frames or images from the same scene), or any setting where the i.i.d. batch assumption does not hold can push these regularizers toward the wrong target. This couples the learning signal to batch composition in a way that the underlying representation problem need not depend on.

\paragraph{Cropping.} Random cropping~\cite{oquab2023dinov2, caron2021emerging} is often treated as a default, but it constitutes a strong inductive bias. It assumes that two crops of the same image should produce similar representations, which in turn assumes both crops contain the same semantic content. On object-centric datasets such as ImageNet this generally holds; on cluttered or scene-level images, however, two crops may capture entirely different objects, and the model is then trained to treat them as equivalent. This can bias representations toward whatever survives random cropping---typically the dominant object---and away from full-scene or denser understanding.

\begin{figure}[t]
    \centering \small
    \includegraphics[width=1\linewidth]{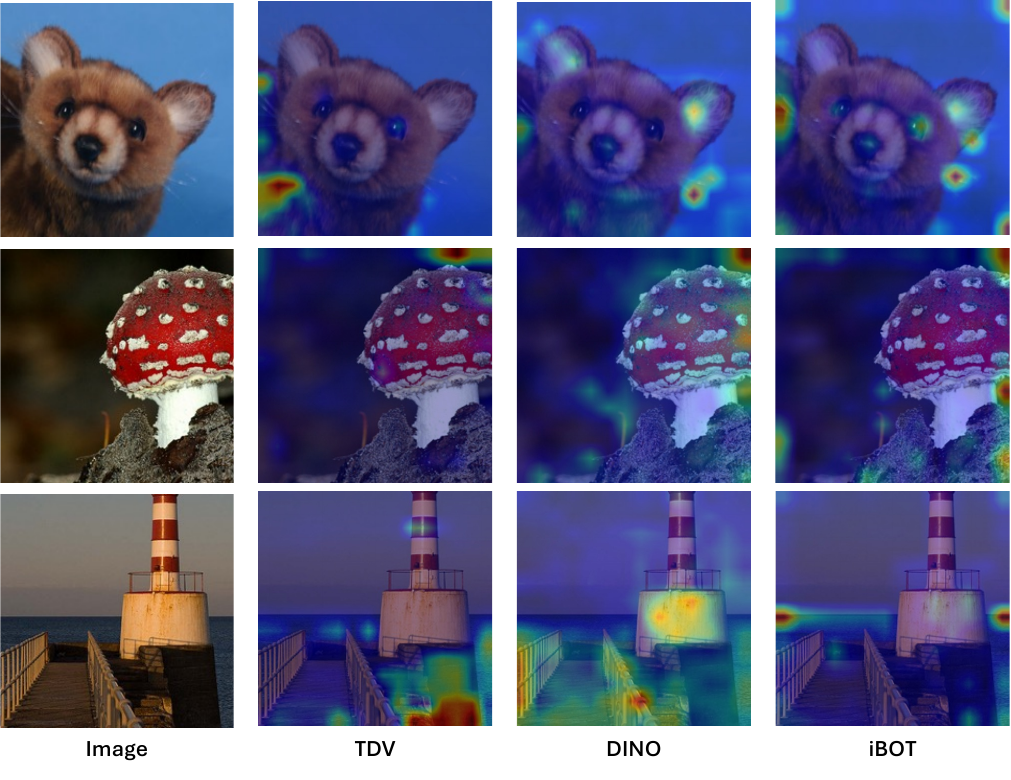}
    \caption{\textbf{Attention Maps Reflect Pre-Training Objectives.} We visualize self-attention maps for the \texttt{[CLS]} token of ViT-B models pre-trained on SSv2 for three ImageNet images (left). Warmer colors indicate higher attention. TDV does not use a dominant \texttt{[CLS]}-token objective during pre-training, hence its \texttt{[CLS]} attention is less object-focused than that of DINO and iBOT.}
    \label{fig:attention_visualization}
\end{figure}


\subsection{Broader Philosophical Intuitions} 

We argue that the ideal inductive bias is the weakest one that still enables learning within a single lifetime of experience---strong enough to bootstrap learning from finite data, but weak enough to not bottleneck what can ultimately be learned. This mirrors biological evolution, where more capable species hardcode less behavior into their genome and instead learn from experience~\cite{tinbergen2020study, reader2002social, gomez2024evolution}: evolution has converged on the weakest priors that still permit survival-level learning within a lifetime. 

Under this minimal-prior view, TDV can be understood as simply performing compression---the motion encoder captures only the change between frames, and the frame encoder captures only what is needed to predict the next frame given that change. No assumptions about augmentation invariances, negative pairs, or pixel-level fidelity are imposed. The model is simply compressing temporal experience into representations sufficient to predict the future. We view this as the minimal assumption necessary for representation learning.

Another perspective on {\pa} is that it is learning a (small) latent action world model~\cite{ha2018world, bruce2024genie}, where the motion encoder captures temporal differences (latent actions), that cause the future frame to be predictable.

\subsection{Neuroscientific Intuitions}
It happens that \pa{} maps reasonably well onto several established theories of biological vision.
The MSE objective---predicting the next-frame embedding from the current one---can be viewed as an instance of \emph{predictive coding}, where the cortex learns by minimizing errors between predicted and observed sensory input~\citep{rao1999predictive, friston2010free, bastos2012canonical}.
The factorization into a frame encoder and a motion encoder operating on $\Delta x$ also loosely mirrors the dorsal/ventral dissociation in primate visual cortex~\citep{Ungerleider1982TwoCV, goodale1992separate}, where the dorsal stream is fed by the magnocellular pathway, itself selective for motion and temporal change.
Most directly, our additive composition $\hat{z}_{t+1} = z_t + \Delta z_t$ echoes the \emph{temporal straightening} hypothesis~\citep{henaff2019perceptual, henaff2021primary}, which posits that the visual system maps video onto straighter latent trajectories so that future states can be predicted by near-linear extrapolation---precisely the structure \pa{} imposes.
While these connections did not directly motivate \pa{}, we find it encouraging that its design choices align with mechanisms already studied in biological vision.
\section{Additional Experiments}
We report results for semantic benchmarking in Table~\ref{tab:semantic_performance}, demonstrating how {\pa} currently lags behind SOTA recipes at semantic representation. This is likely due to a lack of strong inductive biases.

\label{sec:additional_experiments}

\begin{table}[!t]
    \centering
    \caption{\textbf{TDV Lags Behind DINO and iBOT on Semantic Evaluations.} We evaluate ImageNet‑1k KNN Top‑1 accuracy and SSV2 action‑recognition Top‑1 
    accuracy for DINO and iBOT with all augmentations enabled compared to 
    {\pa} (all trained on the Something-Something V2 video dataset). {\pa}, 
    being trained without any strong inductive biases, does not achieve the same 
    semantic performance as iBOT or DINO.}
    \begin{tabular}{l l c c c}
    \toprule
    & & \multicolumn{2}{c}{\textbf{ImageNet Classification}} & \multicolumn{1}{c}{\textbf{SSv2 Action Recognition}} \\
    \cmidrule(lr){3-4} \cmidrule(lr){5-5}
    Method & Architecture & KNN (Top-5) $\uparrow$ & Linear (Top-5) $\uparrow$ & Linear (Top-5) $\uparrow$ \\
    \midrule
    iBOT    & ViT-S      & 33.46 & 41.29 & 20.30 \\
    DINO    & ViT-S      & 34.81 & 44.19 & 19.50 \\
    {\pa}   & ViT-S      & 14.74 & 17.52 & 10.10 \\
    \midrule
    iBOT    & ViT-B      & 38.75 & 46.10 & 21.60 \\
    DINO    & ViT-B      & 40.89 & 49.38 & 20.50 \\
    {\pa}   & ViT-B      & 17.05 & 16.14 & 10.10 \\
    \bottomrule
\end{tabular}
    \label{tab:semantic_performance}
\end{table}

\subsection{Experiments We Tried that Didn't Work}

\label{sec:negative_results}

Inspired by~\cite{redmon2018yolov3}, we document design choices and training strategies that we explored but found to be ineffective. It's worth noting that because {\pa} doesn't leverage any strong inductive biases, in general, training it with current tools is not easy.

\paragraph{\bf Scaling to larger video datasets.}
We explored pretraining on Ego4D~\cite{grauman2022ego4d} and FineVideo~\cite{Farre2024FineVideo} as alternatives to SSv2, to test if more data would improve representation quality. Ego4D is an egocentric dataset with significant variance in motion where some clips are nearly static while others have fast, erratic camera motion, which we found made the RGB difference signal noisy and difficult for the motion encoder to learn from consistently. FineVideo contains many abrupt scene cuts where the difference between consecutive frames reflects an editing transition rather than natural motion. We preprocessed it into smaller chunks to avoid exposing the model to cross-scene differences, but this substantially reduced the usable data volume. Despite our preprocessed FineVideo having approximately 2$\times$ more video data than SSv2, pretraining on it for the same 200k steps yielded a lower ImageNet KNN Top-5 accuracy (10.75\% vs.\ 17.05\% with SSv2). A good finding was that if you kept training {\pa} on FineVideo for longer, the representation quality keeps consistently improving, reaching equal KNN performance (16.04\%) to SSv2 at around 600,000 steps. This suggests that the data quality and motion coherence also matter for the current {\pa} architecture. Future work can address making {\pa} more robust to noisy data. 

\begin{figure} [t]
    \centering
    \includegraphics[width=1\linewidth]{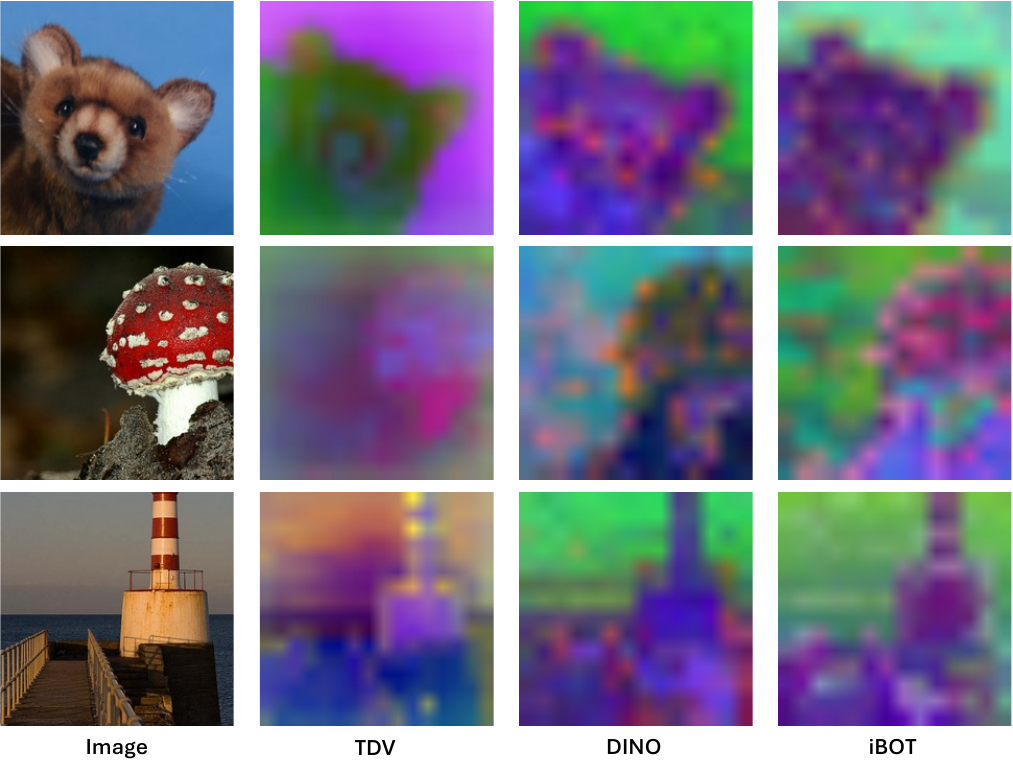}
    \caption{\textbf{PCA Visualization Shows Patch Features Capture Coherent Object Structure.} We show RGB visualizations of the top-3 PCA components of patch-level features from ViT-B models pre-trained on SSv2, for three ImageNet images (left). Each color corresponds to a principal direction of the patch-feature space, so similar colors mark patches with similar representations. TDV produces clean, spatially coherent feature maps that align with object boundaries, often better than DINO and iBOT. This shows that TDV learns strong patch-level representations, consistent with its dense-prediction performance (Tab.~\ref{tab:downstream_flow_stereo}).}
    \label{fig:pca-vis}
\end{figure}

\paragraph{\bf Combining multiple datasets.}
We experimented with mixing SSv2, Kinetics-400~\cite{kay2017kinetics}, and Ego4D in various combinations to increase training data diversity. In all cases, representation quality as measured by ImageNet KNN degraded relative to training on individual datasets alone. We attribute this to the heterogeneous motion statistics across datasets: Kinetics clips tend to be short and action-centric, while Ego4D has highly variable motion rates, and mixing these with SSv2 appears to make the motion prediction task less coherent rather than more informative. Future work can search for hyperparameters that generalize better to multiple datasets.

\paragraph{\bf Alternative conditioning mechanisms for the motion encoder.}
The motion encoder in {\pa} is conditioned on the frame encoder's \texttt{[CLS]} + patch tokens via standard cross-attention. We explored replacing this with feature-wise linear modulation (FiLM)~\cite{perez2018film}, AdaLN, AdaLN-Zero~\cite{peebles2023scalable}, and Gated AdaLN as conditioning mechanisms. All four alternatives showed promising KNN accuracy during the first few training epochs---in all cases doing 2x better than standard cross attention---but then plateaued or collapsed after a few epochs of training. The best observed KNN Top-5 values were FiLM: 7.92\%, AdaLN: 8.97\%, AdaLN-Zero: 10.21\%, and Gated AdaLN: 9.91\% vs Cross Attention: 5.4\% (at that epoch). We hypothesize that while these modulation-based approaches provide stronger conditioning signal, they make it easier for the model to find degenerate solutions, directly affecting the training stability. Future work can leverage newer anti-collapse solutions, e.g. SIGReg~\cite{balestriero2025lejepa}.

\paragraph{\bf RGB difference thresholding.}
We experimented with skipping the motion encoder pass for frames where the magnitude of the RGB difference fell below a threshold, with the intention of filtering out near-static frames where the motion signal is near-zero. In practice this did not improve results. We suspect that static frames still contribute useful learning signal by requiring the motion encoder to produce a near-zero delta, which may provide a natural calibration for the prediction objective.

\paragraph{\bf Motion embedding divergence loss.}
We explored adding an auxiliary loss to push consecutive frame encoder embeddings further apart, with the intuition that a larger gap in representation space would give the motion encoder more room to learn a meaningful delta. In practice, this caused the variance of frame representations to explode and destabilized training without improving KNN or downstream performance.

\paragraph{\bf iBOT-style grid masking.}
We attempted to combine {\pa} with iBOT-style~\cite{zhou2021ibot} patch-level masked prediction, to see if adding more augmentations improves spatial performance. The main challenge is in handling masking consistently between the current frame and the RGB difference input: simply masking the same patch positions from both inputs causes collapse for masking rates above 10\%. Below that threshold, performance was not meaningfully improved over the unmasked baseline, suggesting the interaction between masking and the motion objective requires more careful design than a simple direct combination.

\paragraph{\bf DINO-style augmentations.}
We experimented with applying standard DINO augmentations (random resized cropping, color jitter, Gaussian blur, solarization) to both the current and next frame before feeding them to the encoders. This consistently led to collapse. Our interpretation is that applying strong augmentations to consecutive frames changes the relationship between them in ways that are inconsistent with the causal motion objective: if two frames are independently cropped to different spatial regions, the RGB difference no longer reflects actual scene motion, removing the core learning signal. More careful augmentation strategies that preserve the temporal relationship between frames may be possible with additional tuning.

\paragraph{\bf Asymmetric teacher sharpening.}
DINO uses a significantly lower softmax temperature for the teacher than the student, effectively sharpening the teacher's output distribution relative to the student's. We explored applying the same asymmetry to {\pa}'s EMA teacher. This consistently performed worse than using the same temperature for both; the best results were obtained when teacher and student were sharpened identically. We do not have a strong explanation for this difference, but note that {\pa}'s teacher plays a different role than DINO's---it supervises next-frame predictions rather than augmented-view agreement---which may be the reason for the change in dynamics of temperature asymmetry.

\paragraph{\bf Per-epoch teacher reinitialization.}
We experimented with reinitializing the teacher frame encoder from the student's latest checkpoint at the start of each epoch, rather than maintaining a continuous EMA. This was motivated by a concern that a slowly-drifting EMA teacher might provide stale targets as training progresses. In practice, this performed worse than a fixed high EMA momentum, likely because abrupt teacher resets remove the smoothing that EMA provides and introduce instability in the supervision signal.

\paragraph{\bf Using full frames instead of RGB differences.}
We tested feeding the full current frame (rather than the RGB frame difference) to the motion encoder, so that both encoders receive complete image inputs. In this setting, the motion encoder has no strong inductive bias toward encoding temporal change, and in practice the model collapsed---the motion encoder tended to replicate the frame encoder's representations rather than learn complementary motion information.

\paragraph{\bf Smaller motion encoders.}
We ablated the size of the motion encoder across a range of parameter counts. Smaller motion encoders consistently yielded lower ImageNet KNN accuracy, with a roughly monotonic relationship between encoder capacity and representation quality. This suggests the motion encoder needs sufficient capacity to model the full range of temporal changes present in video, and undersizing it creates a bottleneck in the learning signal reaching the frame encoder.

\paragraph{\bf Continued pretraining of existing vision encoders.}
We explored using {\pa} to improve existing pretrained vision encoders (MAE~\cite{he2021masked} and DINOv2~\cite{oquab2023dinov2}) by initializing the frame encoder from their weights and continuing pretraining with small learning rates while jointly training a motion encoder. With the frame encoder unfrozen, this consistently degraded the pretrained representations rather than improving them. However, keeping the pretrained frame encoder \emph{frozen} and training only the motion encoder does work: the motion encoder successfully learns to predict frame embedding differences, recovering approximately 90\% of the embedding delta for MAE and 60\% for DINOv2. This suggests {\pa}'s motion objective is compatible with existing pretrained features as a fixed representation backbone. This could be useful for efficient video encoding.

\section{{\pa} Details}
\label{sec:approach_details}

\subsection{Training Recipe}
\label{sec:training_recipe}

Algorithm~\ref{alg:tdv} summarizes the {\pa} training procedure for a single step. The student frame encoder and motion encoder are updated by gradient descent; the teacher frame encoder is updated only via EMA and receives no gradients. We also summarize the differences between {\pa} and DINO/iBOT in Table~\ref{tab:augmentations}.

\begin{table}[t]
    \centering
    \caption{\textbf{{\pa} Requires No Hand-Crafted Augmentations.} DINO and iBOT both rely on heavy augmentations. {\pa}, by contrast, uses none of these inductive biases, while still avoiding collapse. Rather, {\pa} learns from the natural change between frames in a video.}
    \label{tab:augmentations}
    \vspace{0.5em}
    \begin{tabular}{l c c c}
        \toprule
        Inductive Bias / Augmentation & DINO & iBOT & {\pa} \\
        \midrule
        Multi-crop (global + local)      & \textcolor{green}{\checkmark} & \textcolor{green}{\checkmark} & \textcolor{red}{\texttimes} \\
        Random resized crop              & \textcolor{green}{\checkmark} & \textcolor{green}{\checkmark} & \textcolor{red}{\texttimes} \\
        Random horizontal flip           & \textcolor{green}{\checkmark} & \textcolor{green}{\checkmark} & \textcolor{red}{\texttimes} \\
        Color jitter                     & \textcolor{green}{\checkmark} & \textcolor{green}{\checkmark} & \textcolor{red}{\texttimes} \\
        Gaussian blur                    & \textcolor{green}{\checkmark} & \textcolor{green}{\checkmark} & \textcolor{red}{\texttimes} \\
        Solarization                     & \textcolor{green}{\checkmark} & \textcolor{green}{\checkmark} & \textcolor{red}{\texttimes} \\
        Masked image modeling            & \textcolor{red}{\texttimes}   & \textcolor{green}{\checkmark} & \textcolor{red}{\texttimes} \\
        \midrule
        Temporal frame sampling          & \textcolor{red}{\texttimes}   & \textcolor{red}{\texttimes}   & \textcolor{green}{\checkmark} \\
        \bottomrule
    \end{tabular}
\end{table}

\begin{algorithm}[t]
\caption{{\pa} Training Step}
\label{alg:tdv}
\begin{algorithmic}[1]
\Require Student frame encoder $f_\theta$, motion encoder $m_\phi$,
         teacher frame encoder $\bar{f}$ (EMA of $f_\theta$)
\State Sample consecutive frames $x_t,\, x_{t+1}$ from a video
\State $z_t \gets f_\theta(x_t)$ \Comment{encode current frame}
\State $\Delta x_t \gets x_{t+1} - x_t$ \Comment{RGB difference}
\State $\Delta z_t \gets m_\phi(\Delta x_t \mid z_t)$ \Comment{encode motion, conditioned on $z_t$}
\State $\hat{z}_{t+1} \gets z_t + \Delta z_t$ \Comment{predict next frame representation}
\State $\bar{z}_{t+1} \gets \bar{f}(x_{t+1})$ \Comment{teacher target (no gradient)}
\State $\mathcal{L} \gets \lambda_\text{mse}\,\underbrace{\|\hat{z}_{t+1} - \bar{z}_{t+1}\|_2^2}_{\text{MSE over all tokens}} \;+\; \lambda_\text{dino}\,\underbrace{\text{DINOLoss}(\hat{z}_{t+1},\, \bar{z}_{t+1})}_{\text{cross-entropy on all tokens}}$
\State Update $\theta,\, \phi$ via backpropagation
\State $\bar{\theta} \gets \tau\,\bar{\theta} + (1-\tau)\,\theta$ \Comment{EMA update teacher}
\end{algorithmic}
\end{algorithm}

\subsection{Pretraining Setup}
\label{sec:pretraining_setup}

All models---{\pa}, DINO~\cite{caron2021emerging}, and iBOT~\cite{zhou2021ibot}---are pretrained on the Something-Something V2 (SSv2)~\cite{goyal2017something} dataset. SSv2 consists of approximately 220,000 short egocentric video clips depicting hand-object interactions, making it a practical choice for initial experimentation due to its manageable size and consistent quality of motion. All models are trained using ViT-S and ViT-B~\cite{dosovitskiy2020image} architectures. Each model is trained for around 200{,}000 steps (20 epochs), and we report results from the final checkpoint. DINO and iBOT are pretrained on SSv2 using their respective objectives and augmentations tuned, which enables their maximum performance on SSv2.

\subsection{Hyperparameters}
\label{sec:hyperparameters}

\paragraph{\bf Model and optimization.}
Table~\ref{tab:hparams} lists the hyperparameters used to train all three models. Shared settings---batch size, optimizer, learning rate schedule, and EMA momentum---are kept identical across methods wherever possible to ensure a fair comparison. Other hyperparameters related to customized augmentations and temperature sharpening were kept at their default values to get the best performance from all recipes.

\begin{table}[t]
\centering
\caption{\textbf{Pretraining Hyperparameters.} Hyperparameters used for {\pa}, DINO, and iBOT pretraining on SSv2.}
\label{tab:hparams}
\small
\begin{tabular}{lccc}
\toprule
Hyperparameter & {\pa} & DINO & iBOT \\
\midrule
Architecture            & ViT-S/B        & ViT-S/B       & ViT-S/B \\
Patch size              & 14             & 16            & 16 \\
Epochs                  & 20            & 20            & 20 \\
Batch size (Images)             & 256            & 256           & 256 \\
Optimizer               & AdamW          & AdamW         & AdamW \\
Learning rate           & 1e-4           & 5e-4          & 5e-4 \\
LR schedule             & cosine         & cosine        & cosine \\
Warmup epochs           & 0.5          & 10        & 10 \\
Weight decay            & 0.01       & 0.04      & 0.04 \\
EMA momentum ($\tau$)   & 0.99       & 0.996      & 0.996 \\
Student temperature ($\tau_s$) & 0.1     & 0.04      & 0.04 \\
Teacher temperature ($\tau_t$) & 0.1     & 0.1      & 0.1 \\
Projection head dim     & 32768       & 1024      & 8192 \\
$\lambda_{\text{mse}}$  & 1.5       & ---           & --- \\
$\lambda_{\text{dino}}$ & 1.5       & ---           & --- \\
\bottomrule
\end{tabular}
\end{table}

\paragraph{\bf Data and temporal sampling.}
Table~\ref{tab:data_hparams} lists the data and temporal sampling hyperparameters specific to {\pa}. The key input is the RGB difference $\Delta x_t = x_{t+1} - x_t$, computed between two temporally adjacent frames sampled at a fixed stride. The stride controls how much motion is visible in the differences between images: too small a stride produces near-zero differences for slow-moving scenes, while too large a stride introduces large, incoherent pixel jumps where object positions change so drastically that the difference no longer captures meaningful motion structure.

\begin{table}[t]
\centering
\caption{\textbf{Data Hyperparameters.} Hyperparameters for Data and Temporal Sampling for {\pa} pretraining}
\label{tab:data_hparams}
\small
\begin{tabular}{lc}
\toprule
Hyperparameter & Value \\
\midrule
Dataset                 & SSv2 \\
Input resolution        & 224 $\times$ 224 \\
Frames sampled per clip & 16 \\
Time Between Frames     & 0.25 \\
RGB difference clipping & no \\
Spatial cropping        & center crop only \\
Horizontal flip         & no \\
Color jitter / augmentations & none \\
Masking                 & none \\
\bottomrule
\end{tabular}
\end{table}

\section{Experimental Details}
\label{sec:experimental_details}

\subsection{Philosophical Backing Experiments}
\label{sec:philosophical_exp_details}

To empirically support our argument that weaker assumptions perform better as data scales, we train a series of models on subsets of ImageNet-1k~\cite{russakovsky2015imagenet} and measure how performance varies with both data scale and augmentation strength. We use the DINO~\cite{caron2021emerging} codebase and augment it with iBOT-style~\cite{zhou2021ibot} patch-level masked prediction, varying the masking ratio as a continuous proxy for assumption strength. To prevent collapse under low-augmentation regimes, we retain random resized cropping with two global crops (consistent with DINO's default setup); all other augmentations are disabled. We vary the grid masking ratio across $\{10\%, 30\%, 50\%\}$ and train each of these masking configurations on data subsets of $\{0.1\%, 1\%, 10\%, 100\%\}$ of ImageNet-1k. We evaluate each model using ImageNet KNN Top-1 accuracy and report the results in Figure~\ref{fig:philosophical_assumption_exps}.

\subsection{Semantic Evaluations}
\label{sec:semantic_eval_details}

In addition to the more semantic/temporally focused representation benchmarking conducted in Section~\ref{sec:experimentation}, we conduct experiments aimed at measuring the semantic representations. Because the default {\pa} training recipe does not leverage augmentations, explicit invariances, or any strong inductive biases, it is expected for learned representations to not be very semantic~\cite{assran2023selfsupervised, misra2020self, purushwalkam2020demystifying, von2021self}. Hence, the expected performance on semantic tasks without these inductive biases is not great. We confirm this in Table~\ref{tab:semantic_performance}, where the semantic performance of the default {\pa} recipe lags behind existing models. 

\paragraph{\bf Action recognition.}
We evaluate action recognition on Something-Something V2~\cite{goyal2017something} following the frozen evaluation protocol of V-JEPA~\cite{bardes2023v}: we freeze the pretrained encoder and train a task-specific linear probe on top of the \texttt{[CLS]} token representations extracted from 8 uniformly sampled frames per video. Frame features are concatenated along the temporal dimension before being passed to the probe head. We report the Top-5 accuracy for action recognition on the validation set for SSv2.

\paragraph{\bf KNN on ImageNet.}
To monitor representation quality and detect collapse during pretraining without incurring the cost of full downstream evaluation, we compute an online ImageNet KNN Top-5 accuracy~\cite{oquab2023dinov2} after each training epoch. At each evaluation step, we extract \texttt{[CLS]} token features for all of ImageNet training subset images using the current student encoder and teacher encoders, and perform $k$-nearest-neighbor classification ($k=20$) in feature space. This provides a lightweight signal that correlates well with final representation quality and allows us to detect collapse early, as collapsed representations yield worse than near-chance KNN accuracy. We report accuracy for ViT small and base variants of iBOT, DINO and TDV in Table \ref{tab:semantic_performance} 

\subsection{Downstream Spatial Evaluations}
\label{sec:downstream_eval_details}

\paragraph{\bf Semantic segmentation.}
We evaluate semantic segmentation using UperNet~\cite{xiao2018unified} via the MMSegmentation~\cite{mmseg2020} toolbox on ADE20K~\cite{zhou2017scene} and Cityscapes~\cite{cordts2016cityscapes} under a frozen backbone evaluation protocol: the pretrained encoder weights are frozen and only the UperNet segmentation head is trained. We use the standard UperNet configuration for ViT backbones and train the segmentation head for 320{,}000 steps for all models. We report mIoU and mAcc on the respective validation sets using the checkpoint from the final training step. All baselines use the same configuration.

\paragraph{\bf Optical flow.}
We follow the evaluation protocol of CroCo~\cite{weinzaepfel2023croco} and perform full end-to-end fine-tuning on FlyingChairs~\cite{dosovitskiy2015flownet}, FlyingThings3D~\cite{mayer2016large} and MPI-Sintel~\cite{butler2012naturalistic} training sets, then evaluate on MPI-Sintel~\cite{butler2012naturalistic} clean and final validation sets using endpoint error (EPE, lower is better). For the decoder, we use the two-frame feature fusion module from Midway Networks~\cite{hoang2025midway}, which processes features from two consecutive frames before passing them to a DPT prediction head (standard CroCo setup). This decoder architecture is identical for all three methods ({\pa}, DINO, iBOT) to ensure fair comparison. We fine-tune all models for 240 epochs using the default CroCo hyperparameters and report results from the final checkpoint.

Because the TDV architecture naturally contains a motion encoder with features that may contain richer temporal correspondences than frame encoder features alone, we additionally experiment with supplying intermediate representations from {\pa}'s motion encoder as an additional input to the DPT head. Removing the Midway Networks\cite{hoang2025midway} decoder and passing embeddings from intermediate layers of the motion encoder directly to the DPT head result in EPE (clean) 14.53 and EPE (final) 14.52 for the {\pa} base variant and EPE (clean) 14.79 and EPE (final) 14.39 for the {\pa} small variant. While these results are currently worse than using the Midway Networks decoder, careful selection of intermediate layers from frame and motion encoder as input to the DPT and more hyperparameter tuning can improve performance directly. We leave this analysis for future work.

\paragraph{\bf Stereo depth.}
We follow the same CroCo~\cite{weinzaepfel2023croco} evaluation protocol for stereo depth: full end-to-end fine-tuning on the SceneFlow (final)~\cite{mayer2016large} training set, evaluated on the SceneFlow final validation set. We use the same Midway Networks~\cite{hoang2025midway} two-frame decoder and DPT head as in the optical flow evaluation. We fine-tune all models for 32 epochs using default CroCo hyperparameters and report average disparity error and bad pixel rates at 0.5px and 1px thresholds from the final checkpoint.

As with optical flow, we experiment with using intermediate representations from {\pa}'s motion encoder as additional input to the DPT head. This variant results in average disparity error 6.93 for the {\pa} base variant and 7.23 for the {\pa} small variant.

\subsection{Compute Resources}
\label{sec:compute_resources}

\paragraph{Pretraining.} All TDV, DINO, and iBOT pretraining runs were conducted on 2 NVIDIA H100 GPUs (80\,GB GPU memory each). Each pretraining run trains for 20 epochs on SSv2 and takes approximately 48 hours.

\paragraph{Semantic segmentation fine-tuning.} UperNet fine-tuning on ADE20K and Cityscapes was run on a single NVIDIA H100 GPU (80\,GB GPU memory) for 320{,}000 steps, taking approximately 20 hours per run.

\paragraph{Optical flow fine-tuning.} Optical flow fine-tuning on FlyingChairs, FlyingThings3D, and MPI-Sintel was run on 2 NVIDIA H100 GPUs (80\,GB GPU memory each) with BF16 precision enabled in the CroCo codebase, taking approximately 48 hours per run.

\paragraph{Stereo depth fine-tuning.} Stereo depth fine-tuning on SceneFlow was run on the same 2 NVIDIA H100 GPUs (80\,GB GPU memory each) for 32 epochs, taking approximately 16 hours per run.

\end{document}